\pdfoutput=1
\documentclass[11pt]{article}

\usepackage[final]{acl}
\usepackage{times}
\usepackage{latexsym}
\usepackage{graphicx}
\usepackage{bookmark}
\usepackage[breakable]{tcolorbox}

\usepackage[T1]{fontenc}
\usepackage[utf8]{inputenc}

\usepackage{microtype}

\usepackage{inconsolata}

\usepackage{graphicx}

\usepackage{hyperref}
\usepackage{graphicx} % Required for inserting images
\usepackage{amsmath}
\usepackage{amsthm}
\usepackage{amssymb}
\usepackage{bbm}
\usepackage[capitalize,noabbrev]{cleveref}
\usepackage{extarrows}
\usepackage{multirow}
\usepackage{tabularx}
\usepackage{booktabs}
\usepackage{tikz}
\usetikzlibrary{fit, calc}
\usepackage{tcolorbox}
\usepackage{colortbl}
\usepackage{xcolor}
\usepackage{pifont}
\usepackage{float}
\usepackage{appendix}
\usepackage{algorithm}
\usepackage{algorithmic}
\usepackage{cleveref}
\usepackage{enumitem}

\usepackage{xspace}
\usepackage{nicematrix}

\newcommand{\ourmethodhighlight}{\textbf{\underline{G}uided \underline{M}odule-\underline{S}ubstitution (GMS)}\xspace}
\newcommand{\ourmethod}{Guided-Module-Substitution (GMS)\xspace}
\newcommand{\ours}{GMS\xspace}

\newcommand{\revision}[1]{#1}
\newcommand{\move}[1]{#1}

\newtcolorbox{revisionbox}[1]{
    colback=gray!5,
    colframe=gray!50,
    title=#1,
    fonttitle=\bfseries,
    breakable
}

\definecolor{mypurple}{RGB}{201,176,219}
\definecolor{purple}{RGB}{112,48,160}
\definecolor{myyellow}{RGB}{251,230,159}
\definecolor{yellow}{RGB}{242,186,2}

\definecolor{green}{RGB}{48,192,180}
\definecolor{blue}{RGB}{72,116,203}
\definecolor{orange}{RGB}{238,130,47}
\definecolor{gray}{gray}{0.3}
\definecolor{yellowgreen}{RGB}{199,213,67}

\definecolor{lightgray}{gray}{0.9}
\definecolor{darkgray}{gray}{0.8}
\definecolor{lightblue}{RGB}{173, 216, 230}

\definecolor{cellbg}{HTML}{EFEFEF}
\definecolor{cellgood}{HTML}{B3E5FC}

\theoremstyle{plain}

\theoremstyle{definition}

\theoremstyle{remark}

\DeclareMathOperator*{\argmax}{argmax}

\usepackage[textsize=tiny]{todonotes}

\usepackage{amsmath,amsfonts,bm}

\def\eqref#1{equation~\ref{#1}}
\DeclareMathAlphabet{\mathsfit}{\encodingdefault}{\sfdefault}{m}{sl}
\SetMathAlphabet{\mathsfit}{bold}{\encodingdefault}{\sfdefault}{bx}{n}

\definecolor{mygray}{gray}{0.6}

\def\eg{{\em e.g.,}\xspace}
\def\ie{{\em i.e.,}\xspace}

\newcommand{\onion}{\textbf{ONION}\xspace}
\newcommand{\zdef}{\textbf{Z-Def.}\xspace}
\newcommand{\pure}{\textbf{PURE}\xspace}
\newcommand{\abl}{\textbf{ABL}\xspace}
\newcommand{\ties}{\textbf{TIES}\xspace}
\newcommand{\dare}{\textbf{DARE}\xspace}
\newcommand{\wag}{\textbf{WAG}\xspace}

\newcommand{\sst}{\textbf{SST-2}\xspace}
\newcommand{\mnli}{\textbf{MNLI}\xspace}
\newcommand{\olid}{\textbf{OLID}\xspace}
\newcommand{\agnews}{\textbf{AGNews}\xspace}

\title{Cut the Deadwood Out: Backdoor Purification \\ via Guided Module Substitution}

\author{Yao Tong$^{1}$\thanks{Equal contribution.}, Weijun Li$^{2*}$, Xuanli He$^3$, Haolan Zhan$^4$, Qiongkai Xu$^2$\thanks{The corresponding author.}\\
$^1$ National University of Singapore, 
$^2$ Macquarie University\\
$^3$ University College London, 
$^4$ Monash University\\
\tt{\small tongyao@u.nus.edu, weijun.li1@hdr.mq.edu.au} \\ \tt{\small xuanli.he@ucl.ac.uk, haolan.zhan@monash.edu, qiongkai.xu@mq.edu.au}
}
\begin{document}

\maketitle
\begin{abstract}
Model NLP models are commonly trained (or fine-tuned) on datasets from untrusted platforms like HuggingFace, posing significant risks of data poisoning attacks. \revision{A practical yet underexplored challenge arises when such \textit{backdoors are discovered after model deployment}, making retraining-required defenses less desirable due to computational costs and data constraints.
In this work, we propose Guided Module Substitution (\ours), an effective retraining-free method based on guided merging of the victim model with just a single proxy model.} Unlike prior ad-hoc merging defenses, \ours uses a guided trade-off signal between utility and backdoor to selectively replaces modules in the victim model.
\revision{\ours offers four desirable properties: (1) robustness to the choice and trustworthiness of the proxy model, (2) applicability under inaccurate data knowledge, (3) stability across hyperparameters, and (4) transferability across different attacks. Extensive experiments on encoder models and decoder LLMs demonstrate the strong effectiveness of \ours}. \ours significantly outperforms even the strongest defense baseline, particularly against challenging attacks like LWS. The code is available at \url{https://github.com/weijun-l/guided-module-substitution}.

%%%\textbf{TL;DR:} A post-training backdoor defense method that greedily substitutes modules in a backdoor model for those in a proxy model, relaxing the assumptions required by previous approaches on clean datasets and clean proxy models.
\end{abstract}

\section{Introduction}\label{sec:intro}
Modern NLP models are frequently adapted to specific downstream tasks through fine-tuning on custom datasets~\cite{howard2018universal}. In practice, these datasets are often collected from diverse sources, some of which may be unreliable (\eg open repositories like HuggingFace~\cite{lhoest-etal-2021-datasets}).
% ding2023parameter
As a result, models can unknowingly incorporate poisoned data, leading to backdoor vulnerability~\cite{zhang2021trojaning, xu2021targeted}: a backdoored model, trained with poisoned data, behaves normally on clean inputs but misbehaves when exposed to the trigger~\cite{10744415}.

\revision{A critical yet underexplored problem arises when such backdoors are discovered after model deployment. Existing defenses, such as data filtering and retraining~\citep{he2024seep}, pruning followed by fine-tuning~\cite{liu2018fine, zhao2024defense}, or unlearning backdoors based on clean~\cite{li2023reconstructive} or poisoned data~\cite{min2024uncovering, li2021anti,chen2022effective}, typically incur substantial computational costs~\cite{wu2022backdoorbench}--especially for large-scale models or proprietary pipelines. This raises a practical and timely question:
\textbf{\textit{Can we effectively purify a trained backdoored model without retraining?}}}

\revision{Recent work~\citep{arora-etal-2024-heres} has proposed model merging as a cost-efficient and retraining-free defense method for model purification: leveraging the abundance of open-source proxy models online\footnote{For reference, on Hugging Face, there are 74 LLaMA-7B and 262 BERT-base models trained on SST-2 alone, and the pool of proxy models is even larger when considering all datasets within the same task domain.}, this approach merges several proxy models--backdoored or not--with the victim model. While promising, this approach relies on \textbf{ad hoc merging of multiple models} with no principled selection, such as utility and defense performance. Its effectiveness is thus unpredictable and sensitive to the number of models involved. Our later experiments in~\cref{sec:main_results} demonstrate that when only a single proxy model is available, such defenses become ineffective. More seriously, In practice, collecting more homogeneous proxy models also requires more effort and often harm utility more~\cite{zhou2025mergeme,wang2025more}, even though more merged parameters can help neutralize backdoors. We therefore target a more challenging scenario--designing guided merging using only one proxy model.}

\revision{In this work, we propose a simple yet effective alternative:
\textbf{guided merging with a single proxy model}. Rather than blindly merging models, we explicitly guide the merging process using a signal that \textit{approximates} the trade-off between utility and backdoor risk. Intuitively, if retaining a particular module from the victim model contributes significantly more to the attack performance than to task utility, it is likely critical for encoding backdoor features. We thus \textit{progressively identify such modules} and \textit{replace} them with their counterparts from the proxy model, as long as the resulting utility degradation remains limited relative to backdoor mitigation gain. We refer to this method as \ourmethodhighlight. While training-free and more efficient than prior model-merging-based defenses, our method \ours not only retains the robustness of model-merging defenses to proxy model trustworthiness, but is also tolerant of imperfect data knowledge, stable across hyperparameters, and exhibits strong transferability across different attacks.} Further discussion of these desirable properties is in~\cref{sec:ablation}.

The contributions of this work are as follows.

\begin{enumerate}[leftmargin=*, nosep]
    \item \revision{We propose \ours, the first backdoor purification method based on guided merging of a victim model with a single proxy model. This design enables purification without retraining while remaining efficient and practical.} 
    \item Experiments on standard NLP models and \revision{LLMs} (\cref{sec:main_results}) show that \ours consistently surpasses competitive defense baselines, with especially strong performance against the more challenging attacks like LWS and HiddenKiller. %Notably, \ours \textit{significantly outperforms all baselines against the more challenging attacks} like LWS and HiddenKiller, \eg on SST-2 victim model, reducing ASR on LWS to 9.7\%, compared to 58.8\% for the best baseline.
    \item \revision{We further validate four desirable properties of \ours in~\cref{sec:ablation} and~\cref{appendix:diff_poison_ratio,appendix:sensitivity_hyperparameter,appendix:clean_victim,appendix:clean_overall_dataset,appendix:backdoor_proxy_model,appendix:proxy_clean_dataset,appendix:retuning,appendix:examples,appendix:transfer}, namely (1) robustness to proxy model choices, (2) applicability to varied data conditions, (3)  stability across hyperparameters, and (4) transferability across attacks.}
    % As shown in~\cref{tab:transfer_attack}, even the weakest strategy (\eg for defending BadNet) can effectively mitigate other backdoor attacks (\eg LWS, HiddenKiller, InsertSent), making it suitable for scenarios where the victim dataset is inaccessible.
    % \item Our method demonstrates stronger robustness to fine-tuning risks (\cref{tab:retuning} and~\cref{appendix:retuning}),
    % % --fine-tuning a purified model on a small proportion of poisoned data is less likely to reintroduce the backdoor--
    % supporting our claim that module-level editing more effectively disrupts backdoor pathways.
\end{enumerate}

\section{Related works}
\label{sec:literature}
\paragraph{Backdoor attack.}
Backdoor attacks aim to manipulate models to behave normally on benign inputs while exhibiting attacker-controlled behavior when specific triggers are present. They can be categorized into two series~\cite{wu2022backdoorbench}: \textit{(1)~Data-poisoning attacks}, exemplified by BadNets~\cite{gu2017badnets}, involve tampering with a subset of training data by adding triggers and altering labels, making them practical in real-world scenarios with minimal attacker assumptions~\cite{turner2019label, carlini2021poisoning, goldblum2022dataset, li2021invisible}. Initially explored in image classification, backdoor attacks have since raised significant security concerns in NLP~\cite{Dai2019ABA,Qi2021TurnTC,Qi2021HiddenKI} with triggers range from misspelled words~\cite{chen2021badnl}, rare words~\cite{yang2021careful, kurita-etal-2020-weight} and syntactic structures~\cite{Qi2021HiddenKI} to text styles~\cite{pan2022hidden}, posing serious challenges for detection. \textit{(2)~Training-control attacks}, on the other hand, assume complete control over the training process and data~\cite{nguyen2021wanet, kurita-etal-2020-weight}, with advances such as layer-wise weight poisoning ensuring backdoor persistence even after fine-tuning~\cite{li2021backdoor}. In this work, we focus on \textit{data-poisoning backdoor attacks in NLP} due to their widespread applicability and practical implications.

\paragraph{Backdoor defense.}
Post-training backdoor defense approaches can be broadly categorized as:
\textit{(1)~Backdoor sample detection followed by training}, which first filters poisoned samples from the dataset and then retrains or fine-tunes the model. These defenses typically rely on the assumption that poisoned samples exhibit distinct characteristics compared to clean (benign) ones, enabling their detection~\cite{qi-etal-2021-onion, yang2021rap, tran2018spectral, he-etal-2023-mitigating}, or attempt to reverse-engineer backdoor triggers to neutralize their impact~\cite{wang2019neural, wang2022universal, wang2023unicorn, xu2024towards}. Afterward, such strategies usually involve unlearning the identified backdoor samples or retraining/fine-tuning on the filtered dataset. However, their effectiveness is often unsatisfactory~\cite{sun2024eliminating, wang2023unicorn, qi-etal-2021-onion}, primarily because accurately identifying all poisoned samples is (increasing) difficult~\cite{wu2022backdoorbench, qi-etal-2021-onion, yang2021rap, tran2018spectral, he-etal-2023-mitigating}, leaving residual backdoor behavior even after retraining or unlearning. Our method, in this regard, compensates for these shortcomings by being more robust to data filtering quality, and thus provides a more reliable alternative to unlearning- or retraining-based defenses.
\textit{(2)~Backdoor model purification:} These methods aim to eliminate backdoor features directly from a well-trained model and are generally regarded as achieving state-of-the-art defense performance~\cite{zhu2023enhancing, zhao2024defense}. Common approaches involve merging parameters with other proxy models~\cite{arora-etal-2024-heres, chen2024neutralizing}, or pruning and fine-tuning using a limited amount of clean data~\cite{wu2021adversarial, min2024towards, liu2018fine, zhao2024defense, zhu2023enhancing}, which may sometimes be assisted by clean proxy models as well~\cite{liu2018fine, zhang2023diffusion}.
Our work advances the second line of research by minimizing the number of auxiliary models to one and by \revision{eliminating the need for additional training and strictly clean datasets}.
%In this paper, we aim to develop effective purification methods.

\paragraph{Model merge.} Model merging methods have emerged as an efficient way to build powerful models by combining existing trained domain-specialized models without requiring extensive retraining~\cite{wortsman2022model,ilharco2022editing,cheng2025whoever}. Most advances focus on developing utility-preserving multi-task models (\eg merging models trained on different tasks like coding and mathematics~\cite{he2025mergebench}), or stronger domain-specific models by combining within-domain models~\cite{wortsman2022model}. Surprisingly, recent work shows that naive weight averaging across multiple models can mitigate backdoors~\cite{arora-etal-2024-heres}. Yet such blind merging is inefficient--requiring unpredictable trial-and-error in model collection, since one cannot know in advance how many models are needed for purification--and often harmful, as merging many models can yield utility drops from parameter interference and redundancy~\cite{zhou2025mergeme,wang2025more}. Thus, our work address these issues by proposing effective and efficient guided merging with a single proxy model. Note that while there are some works studying how model merging systems themselves can be exploited for backdooring, leveraging the fact that merging methods aim to preserve (task) parameters from all models~\cite{wang2025purity,zhang2024badmerging,yuan2025merge}, our objective is fundamentally different: we use merging as a tool to purify a single target model, where only the target model's clean task is relevant and other tasks can all be discarded. Since our main goal is to demonstrate guided merging can act as a defense--and obtaining a single homogeneous proxy is practical--we focus on the standard homogeneous setting. As heterogeneous merging itself remains an open challenge~\cite{xu2024training,du2025adamms}, exploring its use for defense is left to future work.

\paragraph{Safety Localization.} Recent studies suggest that safety-critical behaviors in LLMs are often localized to specific layers or components. For example, jailbreak defenses identify \textit{safety layers} that are critical for aligning harmful queries~\cite{zhao2024defending,ouyang2025layer,zhou2024alignment}; a small set of \textit{safety layers}~\cite{li2024safety} or \textit{safety modules} (\eg attention heads)~\cite{zhou2024role} are crucial for distinguishing malicious from benign inputs; and modifying specific safety-related parameters can further enhance alignment~\cite{wei2024assessing}. Similarly, backdoor research shows that certain layers~\cite{jebreel2023defending} and even neurons or heads~\cite{zhao2024defense} disproportionately influence attack success, enabling defenses through targeted editing or masking. These findings motivate our focus on module- and layer-level guided merging, which balances efficiency with effectiveness.

\section{Preliminary} \label{sec:preliminary}
% Modern neural network architectures can be viewed as compositions of \textit{functional blocks}, where each block serves as a reusable unit performing a well-defined computational operation integral to the model's function, such as Transformer blocks in Transformer-based models, and blocks composed of convolutional and pooling layers (\eg VGG, ResNet, or DenseNet blocks) in modern CNNs \xuanli{ref? probably cite CNN and Transformer paper}. Each functional block comprises \textit{representative functional modules}, which are distinct parameters responsible for executing the block's specific computations.
% For example, in Transformer-based models, these include the query, key, value, and output projection weights in the attention mechanism, as well as the feed-forward transformation and final projection weights in the feed-forward network. 
Modern neural network (NN) architectures can be viewed as layer-wise compositions of \textit{functional blocks}, where each block serves as a reusable unit performing a well-defined computational operation. Hereby, we define two aspects to systematically describe generic NN architectures:
\begin{enumerate}[leftmargin=*, nosep]
    \item \textbf{Layer Set} ($\mathcal{L}$) is the set of block indices that captures the depth of the model. 
    \item \textbf{Module Set} ($\mathcal{M}$) is the set of representative functional modules within each block. 
\end{enumerate}

\paragraph{Transformer blocks.}
In this paper, we use Transformer-based architecture~\cite{vaswani2017attention} as the testbed, given its widespread attention and implementation in NLP.

For each Transformer block, the input will be processed through Attention and Feed-Forward Network (FFN) modules. An attention module contains four components: \(W_Q\), \(W_K\), \(W_V\), and \(W_O\), projecting the input into \textit{query}, \textit{key}, \textit{value}, and the final \textit{output} representation, respectively. The following FFN contains two components: \(W_F\) for the forward layer and \(W_P\) for the projection layer, connected by a non-linear activation function.
% Let $M$ be a Transformer-based language model (LM) model with $L$ Transformer blocks. The output of the \( l \)-th block is denoted as \( H^l\) for \(l \in \{1, 2, \ldots, L\}\). Each Transformer block consists of a multi-head self-attention (MHSA) mechanism\footnote{For simplicity, we do not differentiate between masked and unmasked MHSA in this context and omit discussion of encoder-decoder models.
% However, our methods can naturally extend to various transformer-based architectures.} and a feed-forward network (FFN). In the computation of the \(l\)-th block, the input \(H^{l-1}\) is first projected into \textit{query}, \textit{key}, and \textit{value} matrices using the learned weights \(W_Q\), \(W_K\), and \(W_V\). Attention scores are then computed across multiple heads based on these projections. The outputs of all attention heads are concatenated and projected using \(W_O\), producing the final output \(\text{Attn}^l\) of the attention mechanism. The feed-forward network (\(\text{FFN}\)) is a two-layer fully connected network with learned weight matrices \(W_F\) and \(W_P\) for the first and second layers, respectively.

In total, each Transformer block contains six key functional modules with parameters \( W_Q, W_K, W_V, W_O, W_F, \) and \( W_P \). For simplicity, we adopt notations \( Q, K, V, O, F, \) and \( P \) to refer to these weight matrices throughout the paper. Accordingly, we \revision{\textbf{\textit{define the module set for Transformer models $\mathcal{M}=\{Q, K, V, O, F, P \}$ and layer set $\mathcal{L}=\{1, 2,\ldots, L \}$}}}.

% In contrast, the post-norm transformer block is formulated as:
% \begin{align}
% \label{eq:transformer-block-mhsa-postnorm}
%     H^l &= \text{LN}\left(\text{FFN}(\text{LN}(O^l + H^{l-1})) + \text{LN}(O^l + H^{l-1})\right), \\
%     O^l &= \text{MHSA}(H^{l-1}).
% \end{align}

% \paragraph{Evaluation metrics}
% Following previous literature~\citep{qi-etal-2021-onion}, we adopt two metrics: (1) Clean Accuracy (CACC) measures the prediction accuracy on clean samples, and (2) Attack Success Rate (ASR) measures the attack accuracy on poisoned test set. A higher CACC reflects better utility, while lower ASR indicates stronger defense.

\begin{figure*}
    \centering
    \includegraphics[width=0.98\linewidth]{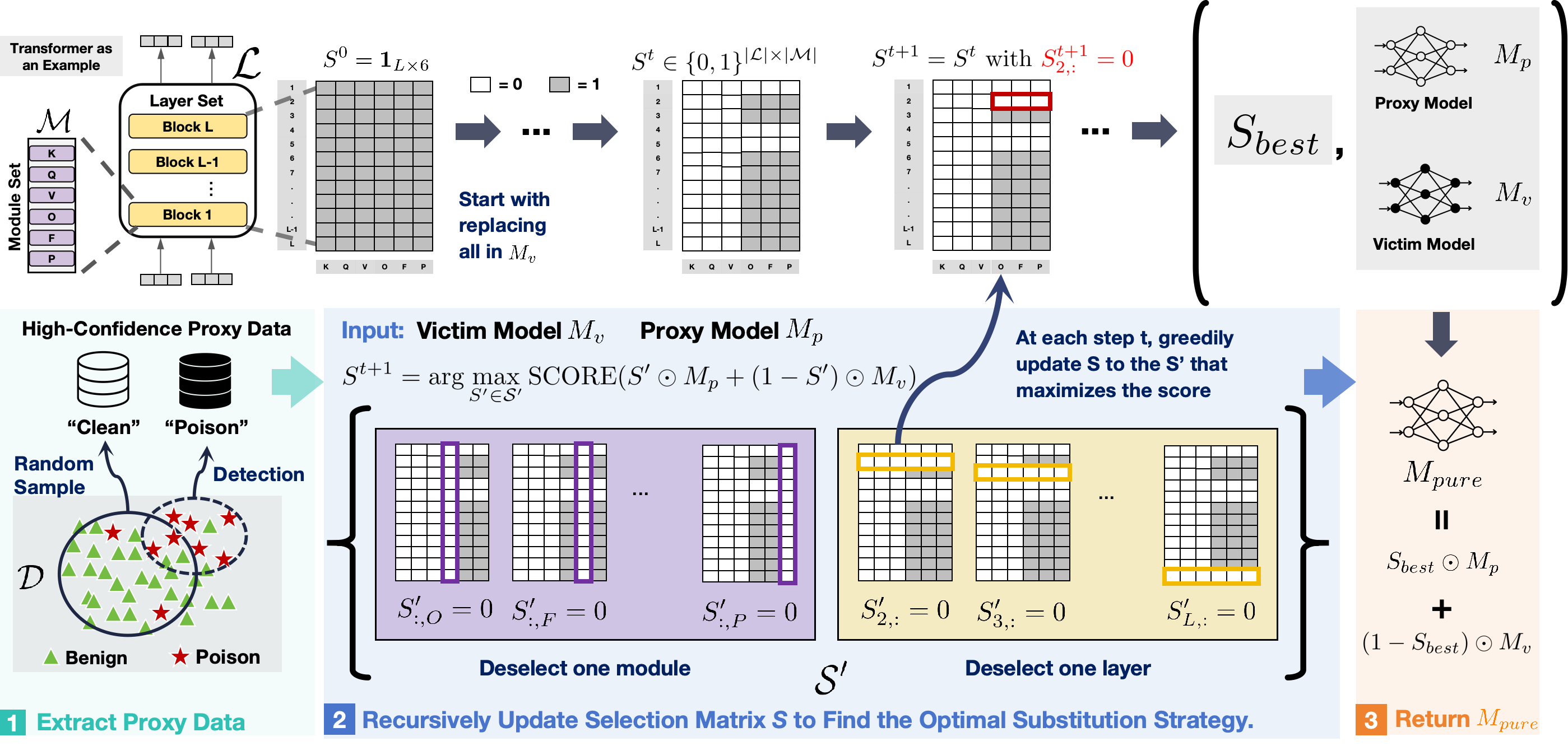}
    \caption{The pipeline of our method. \textcolor{green}{Step 1: Extract two small proxy datasets} used for computing score in~\cref{eqn:objective}. \textcolor{blue}{Step 2: Iteratively update a substitution matrix \( S \)} to greedily maximize the trade-off score between backdoor removal and utility preservation. \textcolor{orange}{Step 3: Return purified model \( M_{pure} \) corresponding to the best substitution matrix \( S_{best} \)}. For more details, refer to~\cref{alg:greedy_search}.}
    \label{fig:pipeline}
    \vspace{-10pt}
\end{figure*}

\section{Methodology}\label{sec:method}
In~\cref{sec:problem_setting}, we introduce the problem setting, and we present our methods in~\cref{sec:our_method}.
% then present the objective function optimized to identify the best parameter substitution strategy and finally describe the complete greedy search pipeline.

\subsection{Problem setting}\label{sec:problem_setting}
% \paragraph{Threat Model.}
% We consider the scenario of data poisoning-based backdoor attacks, where an attacker releases a poisoned dataset $\mathcal{D}$ to embed backdoors into any model trained on this dataset. Starting with a clean dataset $\mathcal{D}_{\text{cl}} = \{(x_i, y_i)\}_{i=1}^{|\mathcal{D}|} \subseteq \mathcal{X} \times \mathcal{Y}$, where $\mathcal{X}$ and $\mathcal{Y}$ denote the input and label spaces, respectively, the attacker selects a subset $\mathcal{D}_{\text{sub}} \subset \mathcal{D}_{\text{cl}}$, and then modified each sample $(x, y) \in \mathcal{D}_{\text{sub}}$ into a poisoned sample $(\tilde{x}, y_t)$ where $\tilde{x} = g(x, \Delta)$, the pre-defined trigger is $\Delta$, $g$ is a generation function, and $y_t$ is the target label. The poisoned subset is $\mathcal{D}_{\text{poi}} = \{(\tilde{x}, y_t) \mid (x, y) \in \mathcal{D}_{\text{sub}}\}$. Finally, the attackers publish the poisoned dataset constructed as $\mathcal{D} = (\mathcal{D}_{\text{cl}} \setminus \mathcal{D}_{\text{sub}}) \cup \mathcal{D}_{\text{poi}}$.

\paragraph{Defense Setting.}
% The defender downloads a victim model \(M_v\) from unauthenticated sources that is likely compromised by a backdoor, along with (a partial snapshot of) its training dataset \(\mathcal{D}\), which is susceptible to poisoning.
% While the defender has access to the dataset \(\mathcal{D}\), they are unaware of the specific backdoor pattern embedded within it. 
% The defender downloads a dataset \(\mathcal{D}\) from unauthenticated sources, which may be vulnerable to poisoning. Consequently, a victim model \(M_v\) trained on it is likely compromised by a backdoor. 
% The defender receives a (potentially) backdoored victim model \(M_v\) trained on an unauthenticated dataset \(\mathcal{D}\). 
% While the defender has access to the dataset \(\mathcal{D}\), they are \textit{unaware} of the specific backdoor pattern embedded within it. 
% Consistent with current mainstream fine-tuning practices, the victim model \(M_v\) is fine-tuned from a benign pretrained model \(M_p\). 
\revision{The defender (\eg a model vendor) has trained a potentially backdoored victim model $M_v$ on an unauthenticated dataset $\mathcal{D}$, which may contain poisoned samples.
Following the standard setting~\cite{li2021anti}, we assume the defender has full control over the training process but lacks prior knowledge of the backdoor patterns, their proportion, or their distribution within $\mathcal{D}$. If an isolation method is employed, it may only identify a subset of the poisoned examples, as perfect detection is not guaranteed.}
The defender has white-box access to the trained victim model $M_v$, as well as access to a proxy model $M_p$ trained on similar downstream tasks.
% The proxy model \(M_p\) does NOT need to be benign.

\subsection{Our method}\label{sec:our_method}
% \revision{Following~\cite{arora-etal-2024-heres}, we aim to merge the victim model with a proxy model. Notice that, unlike standard model-merging, our goal is not to preserve the utility of both models' task. This allows us to adopt a more aggressive substitution strategy to remove backdoor: we directly replace backdoor-encoded modules in the victim model \(M_v\) with counterparts from the proxy model \(M_p\).}

\revision{We aim to combine a victim model with a proxy model. Unlike traditional model merging~\cite{arora-etal-2024-heres}, which prioritizes utility across all models' tasks, our method targets backdoor removal and requires preserving only the clean task performance of the target model. This enables a more targeted substitution strategy: directly replacing backdoor-encoded modules in the victim model \(M_v\) with counterparts from the proxy model \(M_p\).} Specifically, we operate at the module- and layer-level, inspired by recent findings that malicious behaviors in LLMs are often localized to specific safety layers and safety modules (discussed in \cref{sec:literature}). While not the only possible granularity, this level strikes a balance between efficiency and effectiveness, and offers greater robustness and transferability than neuron-level edits, as discussed in~\cref{sec:our_method_algo} and~\cref{sec:ablation}.

Our main algorithm, \ourmethodhighlight, is detailed in~\cref{sec:our_method_algo}. \cref{sec:objective} defines the objective optimized by \ours, and \cref{sec:aux_datasets} provides an example of constructing proxy datasets. %objective 
% \xqk{Need to add some words to guide the readers to focus on algorithm in 4.2.3 (Please also add intuition) and 4.2.1 and 4.2.2 are explaining the objective of the selective algorithm to optimize and the method to construct proxy datasets for evaluating the performance.}
% The victim model demonstrates strong utility on the main task but may contain backdoors, while the proxy model (\eg a model trained on similar tasks) has lower ASR on victim attack but might exhibit lower task utility. 

\subsubsection{Objective}\label{sec:objective} 
The defender's objective is to identify a purified model \( M \) within the model space $\mathbb M$, defined by all possible substitution strategies over the layer set $\mathcal{L}$ and module set $\mathcal{M}$, that effectively eliminates the backdoor while preserving task utility:
\begin{equation}
     \argmax_{M} \, (1-\alpha) \cdot \Delta_\text{asr}(M) %\nonumber 
     + \alpha \cdot (1-\Delta_\text{acc}(M)) \label{eqn:objective}
\end{equation}
where \(\Delta_\text{asr}(M)\)is a score that reflects the extent of \textit{backdoor removal} and \(\Delta_\text{acc}(M)\) assesses the \textit{reduction in task utility}. \move{Specifically, if given access to clean and poisoned samples, we can evaluate the attack success rate (ASR) on poisoned data and the clean accuracy (ACC) on benign inputs. The improvement in backdoor defense and degradation in utility for a given model \( M \) relative to the original victim model $M_v$ are defined as:
\begin{align}
    \Delta_\text{asr}(M) & = s_{asr}(M_v) - s_{asr}(M), \label{eqn:asr} \\ 
    \Delta_\text{acc}(M) & = s_{acc}(M_v) - s_{acc}(M). \label{eqn:acc}
\end{align}}
The hyperparameter \(\alpha \in [0, 1]\) controls the trade-off between backdoor removal and task utility preservation. 
A larger $\alpha$ favors utility, while a smaller $\alpha$ prioritizes backdoor defense. 
% \revision{However, in practice, the ground-truth ACC and ASR cannot be accurately evaluated, as the backdoor pattern is unknown. Following the standard pipeline~\cite{li2021anti}, we assume that a data detection or isolation method may be available. In~\cref{sec:aux_datasets}, we briefly introduce the heuristics used in this paper to approximate these two signals. We also highlight the favorable property of our method: it remains effective under relaxed data knowledge assumptions.}
% Here, \( s_{acc}(M) \)  represents the prediction accuracy of 
% \( M \) on \(\mathcal{D}_{clean}\). \( s_{asr}(M) \) denotes \( M \)'s (attack) accuracy on \(\mathcal{D}_{poison}\), where samples are likely label-flipped poisoned data. Similarly, \( s_{acc}(M_v) \) and \( s_{asr}(M_v) \) denote the original victim model \( M_v \)'s accuracy on \(\mathcal{D}_{clean}\) and \(\mathcal{D}_{poison}\). In this context, \(\Delta_\text{asr}(M)\) quantifies the decrease in attack accuracy on poisoned-like samples achieved by a purified model $M$, while \(\Delta_\text{acc}(M)\) reflects the impact in task utility. Plugging them back into the objective (\cref{eqn:objective}) gives us an score.\footnote{An inaccurate \(\mathcal{D}_{poison}\) may lead to a smaller \(\Delta_\text{asr}\), while an impure \(\mathcal{D}_{clean}\) might result in a larger \(\Delta_\text{acc}(M)\). However, since the metrics are always calculated on fixed proxy datasets, these effects can be controlled by adjusting \(\alpha\).} 

\subsubsection{Proxy datasets $\mathcal{D}_{poison}$ and $\mathcal{D}_{clean}$} \label{sec:aux_datasets}
% In this section, we introduce the process of extracting two proxy datasets from the suspected dataset \(\mathcal{D}\): a proxy clean set \(\mathcal{D}_{clean}\), which is identified \textit{with high probability} as clean (\ie they are likely to be clean though not guaranteed to be free from poisoning), and a proxy poison set \(\mathcal{D}_{poison}\), which is identified \textit{with high confidence} as being poisoned. 

To approximate~\cref{eqn:asr,eqn:acc}, we extract two proxy datasets: a proxy clean set $\mathcal{D}_{\text{clean}}$ and a proxy poison set $\mathcal{D}_{\text{poison}}$, without assuming any prior knowledge of the backdoor. We only follow the standard setting to assume access to the (potentially poisoned) training set $\mathcal{D}$, or to a significantly smaller subset (\eg 1\%) of it (\eg in user-reporting scenarios).

In literature, various methods can be used to construct proxy datasets under such settings, such as sample diagnosis~\cite{gao2019strip, huang2022backdoor, guo2023scale} and trigger inversion~\cite{wang2019neural, wang2022universal, wang2023unicorn, xu2024towards}. For simplicity, we adopt the following heuristics in this paper:
we construct $\mathcal{D}_{\text{clean}}$ via \textbf{random sampling}, assuming that poisoned samples do not dominate the dataset;
and we construct $\mathcal{D}_{\text{poison}}$ using a \textbf{naive confidence-based heuristic}, based on the observation that poisoned examples often yield higher output confidence scores on backdoored models~\cite{li2021anti, swayamdipta2020dataset}.

% . To construct the proxy poison set \(\mathcal{D}_{poison}\), we adopt an existing backdoor sample detection method from SEEP~\cite{he2024seep}, which leverages an observation that backdoored models tend to exhibit over-confidence on poisoned samples~\cite{li2021anti}. SEEP utilizes training dynamics to identify anomalous samples in the training corpus, which are proven to have a high probability of being backdoor samples. In our example, we only focus on the most confident samples extracted (\ie seed set).
% introduces a score to distinguish typical poisoned data from benign samples: the mean inverse confidence \(\frac{1}{E} \sum_{e=1}^{E} \frac{1}{1 - p(y_i | x_i; \theta_e)}\), where \(E\) is the total number of training epochs, and \(p(y_i | x_i; \theta_e)\) represents the predicted probability of the gold label \(y_i\) for input \(x_i\) under the model parameters \(\theta_e\) at epoch \(e\). Poisoned data are therefore identified as those with larger inverse confidence scores, as the over-confidence characteristic of backdoored models results in \( p(y_i | x_i; \theta_e) \) being closer to 1. 

\subsubsection{Greedy search for purified model $M_{pure}$}\label{sec:our_method_algo}
Using the extracted proxy datasets $\mathcal{D}_{\text{poison}}$ and $\mathcal{D}_{\text{clean}}$, we compute (potentially inaccurate) estimates of~\cref{eqn:asr,eqn:acc}\footnote{An imperfect $\mathcal{D}_{\text{poison}}$ will lead to an underestimated $\Delta_\text{asr}(M)$, while an impure $\mathcal{D}_{\text{clean}}$ results in an overestimated $\Delta_\text{acc}(M)$. However, what matters is only the \textit{relative scaling} between these two after weighting by $\alpha$ (\cref{appendix:proof_random_sampling}).}and plug into the objective~\cref{eqn:objective} to get the scoring function that guides the substitution process.

% We define below two metrics:
% \begin{align}
%     \Delta_\text{asr}(M) & = s_{asr}(M_v) - s_{asr}(M), \label{eqn:asr} \\ 
%     \Delta_\text{acc}(M) & = s_{acc}(M_v) - s_{acc}(M). \label{eqn:acc}
% \end{align}
% Here, \( s_{acc}(M) \)  represents the prediction accuracy of 
% \( M \) on \(\mathcal{D}_{clean}\). \( s_{asr}(M) \) denotes \( M \)'s (attack) accuracy on \(\mathcal{D}_{poison}\), where samples are likely label-flipped poisoned data. Similarly, \( s_{acc}(M_v) \) and \( s_{asr}(M_v) \) denote the original victim model \( M_v \)'s accuracy on \(\mathcal{D}_{clean}\) and \(\mathcal{D}_{poison}\). In this context, \(\Delta_\text{asr}(M)\) quantifies the decrease in attack accuracy on poisoned-like samples achieved by a purified model $M$, while \(\Delta_\text{acc}(M)\) reflects the impact in task utility. 

\revision{While our goal is to purify \( M_v \) by replacing its modules across \( L \) layers, an exhaustive search is computationally impractical due to its exponential complexity (\eg \( 2^{6 \times 12} \) for a \textit{BERT-base} model~\citep{devlin-etal-2019-bert}). Moreover, considering individual parameters as functional units for editing often falls short in breaking backdoor connections (thereby leading prior purification works to heavily rely on clean fine-tuning to restore clean features~\cite{liu2018fine}), we propose a greedy algorithm that transforms the problem into a feasible search--balancing granularity and scalability by iteratively identifying and replacing the most critical module, maximizing the objective score to derive the optimal purified model \( M_{pure} \).}
%We propose a greedy algorithm to iteratively refine \( M_v \), maximizing this score to obtain the optimal purified model \( M_{pure} \).

\begin{algorithm}[!t]
% \footnotesize
\small
\caption{\ourmethod}
\label{alg:greedy_search}
\begin{algorithmic}[1]
\STATE \textbf{Input:} Victim model $M_v$, proxy model $M_p$, module set $\mathcal{M}$, layer set $\mathcal{L}$, proxy datasets $\mathcal{D}_{clean}$ and $\mathcal{D}_{poison}$
\STATE \textbf{Output:} Purified model $M_{pure}$

\STATE Initialize $S \gets \mathbf{1}_{|\mathcal{M}| \times |\mathcal{L}|}$, $c_{best}\gets -\infty$, $S_{best}\gets \varnothing$
\STATE $(s_{acc}, s_{asr}) \gets$ Evaluate($M_v$)
% \STATE \textbf{\textcolor{gray}{\# Iteratively update selected modules and layers}}
\STATE \textbf{\textcolor{mygray}{\# Iteratively Update Selected }\textcolor{purple}{Modules }\textcolor{mygray}{ and }\textcolor{yellow}{Layers}:}
\WHILE{$|\mathcal{M}| > 1$ \textbf{or} $|\mathcal{L}| > 1$}
    % \STATE \textbf{\textcolor{gray}{\# Compute Scores for }\textcolor{purple}{Modules }\textcolor{gray}{ and }\textcolor{yellow}{Layers}:}
    \STATE $c^* \gets -\infty$, $S^* \gets \varnothing$
    \FOR{each \textcolor{purple}{$m \in \mathcal{M}$} or \textcolor{yellow}{$l \in \mathcal{L}$}}
        \STATE $S' \gets S$ with $S'_{:,m} = \mathbf{0}_{L\times 1}$ or $S'_{l,:} = \mathbf{0}_{1\times |\mathcal{M}|}$
        \STATE $M_{pure}' \gets$ Replace($M_v$, $M_p$, $S'$)
        \STATE $c \gets$ ComputeScore($M_{pure}, s_{acc}, s_{asr}$)
        \IF{$c > c^*$ }
            \STATE Update $c^* \gets c, S^* \gets S'$
        \ENDIF
    \ENDFOR
    % \STATE \textbf{\textcolor{gray}{\# Iteratively update selected modules and layers}}
    \STATE $S \gets S^*$ and update $\mathcal{M}, \mathcal{L}$ accordingly
    \IF{$c^* > c_{best}$ }
        \STATE Update $c_{best} \gets c^*$, $S_{best} \gets S^*$
    \ENDIF
    \IF{$c_{best}$ not been updated for $T$ iterations}
        \STATE \textbf{break}
    \ENDIF
\ENDWHILE

\RETURN $M_{pure} = \text{Replace}(M_v, M_p, S_{best})$ \label{alg_line:return}
\end{algorithmic}
\end{algorithm}
\paragraph{Purifying $M_v$ by module substitution.}
% As analyzed in~\cref{sec:intro}, pruning or single-parameter-level editing is insufficient to break the backdoor-connected path. Instead, a more effective strategy is to directly replace the parameters of an entire module (\eg $K,Q,V,O,F,P$)) in specific layers. This motivates our proposed substitution defense: 
Given a victim model \( M_v \) (likely to has high ACC and high ASR), we next illustrate how to identify the specific modules and layers that, when substituted, allow the model to maintain strong performance on the clean task (\ie minimizing $\Delta_\text{acc}$) while eliminating the backdoor features (\ie maximizing $\Delta_\text{asr}$).

Our algorithm, as demonstrated in~\cref{fig:pipeline} and~\cref{alg:greedy_search}, tracks the parameters selected for substitution using a module set ($\mathcal{M}$) and a layer set ($\mathcal{L}$), \eg if \( \mathcal{M} = \{O, F, P\} \) and \( \mathcal{L} = \{7, 8, 9\} \), it indicates that the \( O \), \( F \), and \( P \) module parameters in the 7th, 8th, and 9th layers of the victim model \( M_v \) should be replaced with the corresponding parameters from proxy model \( M_p \). For simplicity, we introduce a substitution matrix $S \in \{0,1\}^{|\mathcal{M}| \times |\mathcal{L}|}$ to specify which modules in which layer in the victim model $M_v$ are selected for substitution. Note that while the shape of \(S\) is determined by the \textit{initial} sizes of sets \(\mathcal{M}\) and \(\mathcal{L}\), these two sets are iteratively updated throughout the algorithm. Each row of \( S \) corresponds to a layer in the layer set \(\mathcal{L}\), and each column corresponds to a module in the module set \(\mathcal{M}\). 
The value of \( S[l,m] = 1 \) indicates that the module \( m \in \mathcal{M} \) in layer \( l \in \mathcal{L} \) in the victim model $M_v$ is selected for substitution, and \( S[l,m] = 0 \) otherwise. 
% The value of \( S[i,j] = 1 \) if the module \( \mathcal{M}[j] \) at layer \( \mathcal{L}[i] \) in the victim model $M_v$ is selected for substitution, and \( S[i,j] = 0 \) otherwise. 
Given $S$, we can succinctly represent the parameter substitution as $S \odot M_p + (1-S) \odot M_v$, where $\odot$ denotes the element-wise product.

% \vspace{-10pt}

Since the primary goal is backdoor removal, we start by selecting all modules in \( M_v \) to be replaced with those of \( M_p \), \ie $\mathcal{M}=\{K,Q,V,O,F,P\}$, $\mathcal{L} = \{1, \ldots,L\}$, $S = \mathbf{1}_{|\mathcal{M}| \times |\mathcal{L}|}$. 
% In each iteration, we maximize the score defined in~\cref{eqn:objective} by either removing one layer (affecting all selected modules) from \( \mathcal{L} \) or removing one module (affecting all selected layers) from \( \mathcal{M} \). 
At $t$-th iteration, the algorithm considers two types of updates to \( S^t \): \textcolor{purple}{deselecting a module} from the current module set \( \mathcal{M} \) (affecting all selected layers) or \textcolor{yellow}{deselecting a layer} from the layer set \( \mathcal{L} \) (affecting all selected modules). The update that maximizes the score in~\cref{eqn:objective} is applied to \( S^{t+1} \).
The process continues until one of two conditions is met: (1) 
% the stopping criteria in Line 20 of~\cref{alg:greedy_search}, indicating that 
further iterations will not yield a better strategy (controlled by stopping patience $T$); or (2) no additional layers or modules can be removed, \ie \( |\mathcal{M}| = |\mathcal{L}| = 1 \). Finally, the strategy $S$ that yields a purified model $M_{pure}$ with the highest score is returned. The complexity of our algorithm is a practical quadratic $O(|\mathcal{M}|^2+|\mathcal{L}|^2)$.

% The more detailed procedure is outlined in~\cref{alg:greedy_search_complete} in Appendix.

% \begin{table*}[ht]
% \centering
% \caption{Baseline comparison on T5.}
% \label{tab:noise_performance}
% \begin{tabular}{@{}lcccccc@{}}
% \toprule
% Baselines & qasc (CACC) & qasc\_poison (ASR) & wiki\_qa & wiki\_qa\_poison & quartz & quartz\_poison \\ \midrule
% Noise Adding & 21.9 & 87.5 & 87.3 & 16.4 & 53.4 & 50.5 \\
% Rank perturbation & 9.4 & 87.5 & 90.2 & 12.3 & 47.4 & 47.4 \\
% Projection & 100.0 & 96.9 & 93.8 & 100.0 & 69.5 & 100.0 \\
% \bottomrule
% \end{tabular}
% \end{table*}

% \begin{enumerate}
%     \item performance under different number of data
%     \item filter out backdoor data with highest confidence (seep)/ other methods to fetch poisoning data
%     \item figure
% \end{enumerate}

% \input{latex/sections/sec5_exp}
\section{Experiments}
\subsection{Experimental settings}
\label{sec:exp_setup}
\noindent \textbf{Datasets.} We evaluate our method on four datasets: \textbf{SST-2}~\citep{socher2013recursive}, \textbf{OLID}~\citep{zampieri-etal-2019-predicting}, \textbf{MNLI}~\citep{N18-1101}, and \textbf{AGNEWS}~\citep{zhang2015character}. These datasets cover text classification and natural language inference (NLI) tasks, including binary and multi-class classification scenarios, and are widely used for evaluating text backdoor attacks and defenses~\citep{qi-etal-2021-onion,he-etal-2023-mitigating, gupta-krishna-2023-adversarial, arora-etal-2024-heres}. We adapt the open-source datasets provided by HuggingFace~\citep{lhoest-etal-2021-datasets}. Table~\ref{tab:dataset} are the statistics of these datasets. 
% To train backdoored models, we construct poisoned datasets by injecting triggers into the training split, while the validation split is used to construct a clean and a poison test set respectively. For MNLI, we use a random subset of 100,000 samples to reduce overhead.

\paragraph{Backdoor Attacks.} We study defenses against four prominent types of \textit{data-poisoning} backdoor attacks: (1) \textbf{BadNets}~\citep{kurita-etal-2020-weight}, (2) \textbf{InsertSent}~\citep{Dai2019ABA}, (3) Learnable Word Substitution (\textbf{LWS})~\citep{Qi2021TurnTC}, and (4) \textbf{HiddenKiller}~\citep{Qi2021HiddenKI}. The first two correspond to insertion-based methods, while the latter two involve synonym substitution and syntactic paraphrasing approaches, respectively. \revision{A more recent attack, \textbf{BITE}~\cite{yan20223bite}, is discussed separately in~\cref{appendix:bite} due to its atypical behavior on benign models.}
% \Such attacks inject predefined triggers or patterns into the training data, manipulate labels to a target class, and introduce spurious correlations that result in unintended behaviour during inference.

Follow \citet{arora-etal-2024-heres}, we use rare words \{\textit{"cf", "mn", "bb", "tq", "mb"}~\} as triggers for \textbf{BadNets} and phrases \{\textit{"I watched this movie", "no cross, no crown"}~\} for \textbf{InsertSent}. The poison target labels for each dataset are listed in Table~\ref{tab:dataset}. While our main experiments focus on 20\% poison rate (in line with the literature~\citep{Dai2019ABA,Qi2021HiddenKI,arora-etal-2024-heres}), \ie 20\% of training samples are poisoned, we also evaluate lower poison rates of 10\%, 5\% in~\cref{appendix:diff_poison_ratio}. 
% While our primary focus is on a high poison rate, where 20\% of training samples are poisoned (in line with~\citep{Dai2019ABA,Qi2021HiddenKI,arora-etal-2024-heres}), we also evaluate lower poison rates of 10\%, 5\%. 
% To ensure a fair comparison, we adopt the identical settings of the state-of-the-art text-based model merging defense, WAG~\citep{arora-etal-2024-heres}, including triggers, target labels, and poison rates. Specifically, we use rare words \{\textit{"cf", "mn", "bb", "tq", "mb"}~\} as triggers for \textbf{BadNets} and phrases \{\textit{"I watched this movie", "no cross, no crown"}~\} for \textbf{InsertSent}. The poison target labels for each dataset are listed in Table~\ref{tab:dataset}. While our primary focus is on a high poison rate, where 20\% of training samples are poisoned (in line with~\citep{Dai2019ABA,Qi2021HiddenKI,arora-etal-2024-heres}), we also evaluate lower poison rates of 10\%, 5\%. 

\begin{table}[t]
    \centering
    \setlength{\tabcolsep}{3pt} 
    \renewcommand{\arraystretch}{1.0} 
    \scalebox{0.78}{
    \begin{tabular}{cccccc}
        % \toprule
        % \textbf{Dataset} & \textbf{Classes} & \textbf{Train} & \textbf{Clean Test} & \textbf{Poison Test}  & \textbf{Target Class} \\
        % \midrule 
        %  SST-2 & 2 & 67,349 & 872 & 444 & Negative (0) \\
        %  OLID & 2 & 13,240 & 860 & 240 & Not offensive (1)  \\
        %  MNLI & 3 & 100,000 & 400 & 285 & Neutral (1)\\
        %  AGNews & 4 & 120,000 & 7,600 & 5700 & Sports (1) \\
        %  \bottomrule
        \toprule
        \multirow{2}{*}{\textbf{Dataset}} & \multirow{2}{*}{\textbf{Classes}} & \multirow{2}{*}{\textbf{Train}} & \multicolumn{2}{c}{\textbf{Test}} & \multirow{2}{*}{\textbf{Target Class}} \\
        \cmidrule(lr){4-5}
        & & & \textbf{Clean} & \textbf{Poison} & \\
        \midrule
        SST-2 & 2 & 67,349 & 872 & 444 & Negative (0) \\
        OLID & 2 & 13,240 & 860 & 240 & Not offensive (1) \\
        MNLI & 3 & 100,000 & 400 & 285 & Neutral (1) \\
        AGNews & 4 & 120,000 & 7,600 & 5,700 & Sports (1) \\
        \bottomrule
    \end{tabular}}
    \caption{The statistics of our evaluated datasets.}
     \label{tab:dataset}
     \vspace{-0.5cm}
\end{table}

\begin{table*}[!t]
\small
\centering
\renewcommand{\arraystretch}{1}
\setlength{\tabcolsep}{5.5pt}
\scalebox{0.9}{
\begin{tabular}{cccccccccc}
\toprule
& & \multicolumn{2}{c}{\textbf{BadNet}} & \multicolumn{2}{c}{\textbf{InsertSent}} & \multicolumn{2}{c}{\textbf{LWS}} & \multicolumn{2}{c}{\textbf{HiddenKiller}} \\
\cmidrule(lr){3-4} \cmidrule(lr){5-6} \cmidrule(lr){7-8} \cmidrule(lr){9-10}
\multirow{-2}{*}{\textbf{Dataset}} & \multirow{-2}{*}{\textbf{Method}} & \textbf{ASR}$\downarrow$ & \textbf{CACC}$\uparrow$ & \textbf{ASR}$\downarrow$ & \textbf{CACC}$\uparrow$ & \textbf{ASR}$\downarrow$ & \textbf{CACC}$\uparrow$ & \textbf{ASR}$\downarrow$ & \textbf{CACC}$\uparrow$ \\
\midrule %%%%%%%%%%%%%%% SST-2  %%%%%%%%%%%%%%%
& \cellcolor{cellbg}Benign & \cellcolor{cellbg}4.1 & \cellcolor{cellbg}95.9 & \cellcolor{cellbg}2.2 & \cellcolor{cellbg}95.9 & \cellcolor{cellbg}12.8 & \cellcolor{cellbg}95.9 & \cellcolor{cellbg}16.5 & \cellcolor{cellbg}95.9 \\
& \cellcolor{cellbg}Victim & \cellcolor{cellbg}100.0 & \cellcolor{cellbg}96.0 & \cellcolor{cellbg}100.0 & \cellcolor{cellbg}96.3 & \cellcolor{cellbg}98.0 & \cellcolor{cellbg}95.4 & \cellcolor{cellbg}96.5 & \cellcolor{cellbg}95.7 \\
& \cellcolor{cellbg}Proxy Model (IMDB) & \cellcolor{cellbg}7.4 & \cellcolor{cellbg}89.1 & \cellcolor{cellbg}4.3 & \cellcolor{cellbg}89.1 & \cellcolor{cellbg}10.8 & \cellcolor{cellbg}89.1 & \cellcolor{cellbg}13.7 & \cellcolor{cellbg}89.1 \\
\cmidrule{2-10}
& ONION & 56.8 & 92.9 & 99.9 & 93.3 & 85.7 & 91.9 & 92.9 & 92.9 \\
& Z-Def. & \cellcolor{cellgood}\textbf{4.6} & \textbf{96.1} & \cellcolor{cellgood}\textbf{1.8} & 95.6 & 97.3 & 95.3 & \cellcolor{cellgood}\textbf{35.7} & 95.4 \\
& PURE & 0.0 & 50.9 & 0.0 & 50.9 & 0 & 50.9 & 0.0 & 50.9 \\
& ABL & 75.0 & 49.2 & 51.7 & 50.8 & 32.3 & 50.3 & 92.9	 & 47.4 \\
\cmidrule{2-10}
& TIES & 99.9 & 95.7 & 100.0 & 95.8 & 93.5 & 95.2 & 88.8 & 95.3 \\
& DARE w/ TIES  & 99.3 & 96.0 & 100.0 & \textbf{96.2} & 96.4 & \textbf{95.7} & 92.7 & \textbf{95.8} \\
& WAG & 84.4 & 94.8 & 60.1 & 95.2  & \cellcolor{cellgood}\textbf{58.8} & 94.8 & 56.2 & 92.5 \\
\cmidrule{2-10}
\multirow{-12}{*}{SST-2} & Ours (\ours) & \cellcolor{cellgood}\textbf{4.5} & 91.6 & \cellcolor{cellgood}\textbf{1.9} & 92.5 & \cellcolor{cellgood}\textbf{9.7} & 91.7 & \cellcolor{cellgood}\textbf{10.4} & 91.2 \\
\midrule   %%%%%%%%%%%%%%% AGNews %%%%%%%%%%%%%%%
& \cellcolor{cellbg}Benign & \cellcolor{cellbg}1.9 & \cellcolor{cellbg}95.4 & \cellcolor{cellbg}0.5 & \cellcolor{cellbg}95.4 & \cellcolor{cellbg}0.5 & \cellcolor{cellbg}95.4 & \cellcolor{cellbg}1.1 & \cellcolor{cellbg}95.4 \\
& \cellcolor{cellbg}Victim & \cellcolor{cellbg}99.9 & \cellcolor{cellbg}95.1 & \cellcolor{cellbg}99.6 & \cellcolor{cellbg}95.3 & \cellcolor{cellbg}99.6 & \cellcolor{cellbg}94.5 & \cellcolor{cellbg}100.0 & \cellcolor{cellbg}95.1 \\
& \cellcolor{cellbg}Proxy Model (BBCNews) & \cellcolor{cellbg}1.5 & \cellcolor{cellbg}70.2 & \cellcolor{cellbg}1.7 & \cellcolor{cellbg}70.2 & \cellcolor{cellbg}1.8 & \cellcolor{cellbg}70.2 & \cellcolor{cellbg}3.4 & \cellcolor{cellbg}70.2 \\
\cmidrule{2-10}
& ONION & 59.4 & 94.8 & 97.8 & 95.1 & 84.8 & 94.5 & 99.6 & 94.7 \\
& Z-Def. & \cellcolor{cellgood}\textbf{1.6} & \textbf{95.3} & \cellcolor{cellgood}\textbf{0.4}& 95.3 & 97.9 & \textbf{96.1} & 100.0 & 95.0 \\
& PURE & 2.8 & 86.3 & 3.0 & 85.4 & \cellcolor{cellgood}\textbf{6.5} & 85.0 & \cellcolor{cellgood}\textbf{9.4} & 84.6 \\
& ABL & 50.2 & 55.0 & 48.8 & 54.8 & 100.0 & 55.0 & 90.2 & 54.9\\
\cmidrule{2-10}
& TIES & 99.9 & 94.6 & 99.6 & 94.4 & 97.7 & 95.8 & 100.0 & 94.4 \\
& DARE w/ TIES & 99.9 &  95.2& 99.6 & \textbf{95.4} & 97.8 & 96.5  & 100.0 & \textbf{95.2} \\
& WAG & 92.7 & 94.1 & 97.8 & 94.4 & 78.0 & 93.9 & 90.9 & 94.3 \\
\cmidrule{2-10}
\multirow{-12}{*}{AGNews} & Ours (\ours) & \cellcolor{cellgood}\textbf{2.5} & 91.0 & \cellcolor{cellgood}\textbf{2.4} & 92.6 & \cellcolor{cellgood}\textbf{3.2} & 91.7 & \cellcolor{cellgood}\textbf{6.5} & 90.4 \\
\bottomrule
\end{tabular}
}
\caption{(\textbf{\textit{Partial}}) Performance of our method on two datasets compared to baselines under various backdoor attacks on the RoBERTa-large model, with each value averaged over three seeds. We highlight the top-2 lowest ASR results in \colorbox{cellgood}{\textbf{blue}} cells, and the highest CACC results in \textbf{bold}. Results for other datasets are in~\cref{tab:avg_roberta_all}.}
\label{tab:avg_roberta}
\vspace{-5pt}
\end{table*}
\paragraph{Defense Baselines.} We choose \revision{\textit{seven}} renowned defenses as main baselines, including \textit{two} data-wise detection methods: (1) \onion~\citep{qi-etal-2021-onion} and (2) \zdef~\citep{he-etal-2023-mitigating}, and \textit{four} model-wise purification methods: (3) \pure~\citep{zhao2024defense}, \revision{(4) \abl~\citep{li2021anti}}, (5) \ties~\citep{yadav2024ties}, (6) \dare~\citep{yu2024language}, and (7) \wag~\citep{arora-etal-2024-heres}.

Among these, ONION and Z-Def. detect and remove triggers from datasets. ONION identifies outlier words (potential triggers) using language models, \eg GPT-2~\citep{radford2019language}, while Z-Def. detects spurious correlations between tokens and labels. 
% In contrast, model-wise purification involves pruning backdoor-related parameters or merging the victim model with others to sanitize it. For pruning-based defenses, we include PURE, which purifies the victim model using attention head pruning and normalization techniques. For model merging, TIES and DARE demonstrate excellent performance in improving model utility, while WAG specifically focuses on defending against backdoor attacks.
In contrast, (2)-(7) are model purification methods. PURE purifies the victim model using attention head pruning and normalization techniques\revision{, and ABL introduces a robust anti-backdoor training framework that first isolates, then unlearns backdoor associations}. TIES, DARE and WAG are mode-merging baselines. We also report three additional baselines in~\cref{appendix:more_baselines}: two model purification methods, \textbf{Fine-mixing}~\citep{zhang2022fine} and \textbf{Fine-purifying}~\citep{zhang2023diffusion}, and one data filtering method, \textbf{SEEP}~\citep{he2024seep}, followed by retraining.
% , while WAG specifically focuses on defending against backdoor attacks.
% These methods typically assume access to different resources, including: i) poisoned dataset (\eg Z-Def and ONION), ii) clean dataset (\eg TIES, DARE and PURE), iii) additional language models like GPT-2 (\eg ONION), iv) homologous models for merging (\eg TIES, DARE, and WAG). Our method relaxes the requirement for clean data by utilizing only the training data and homologous models to purify the model.

\paragraph{Evaluation Metrics.} Following previous literature~\citep{qi-etal-2021-onion}, we adopt Clean Accuracy (\textbf{CACC}) and Attack Success Rate (\textbf{ASR}) to measure utility and defense performance respectively. CACC is evaluated on a clean test set, where both the samples and labels are ground truth. In contrast, ASR is evaluated on a poisoned test set, where triggers are implanted into each sample, and the labels are flipped to the target class.

% \yao{mentioned in~\cref{sec:preliminary} already}
% Following previous literature~\citep{Dai2019ABA,Qi2021HiddenKI,arora-etal-2024-heres}, we adopt two evaluation metrics: Clean Accuracy (CACC), which measures the prediction accuracy on clean samples, and Attack Success Rate (ASR), which measures the prediction accuracy of poisoned samples classified into the target class. Higher CACC reflects better utility performance, while lower ASR indicates stronger defense performance.

\paragraph{Implementation details.} 
Following~\citet{arora-etal-2024-heres}, we compare all methods on \textit{RoBERTa-large}~\citep{DBLP:journals/corr/abs-1907-11692}, \textit{BERT-base-uncased}~\citep{devlin-etal-2019-bert}, \revision{as well as LLMs including \textit{Llama 2 7B}~\citep{touvron2023llama}, \textit{Mistral 7B}~\citep{jiang2023mistral}, and \textit{Qwen 2.5 7B}~\citep{qwen2.5}, with low-rank adaptation LoRA~\citep{hu2021lora}}. All victim models are fine-tuned on poisoned datasets using the Adam optimizer with no weight decay~\citep{kingma2014adam} and a learning rate of $2 \times 10^{-5}$. All defense baselines are implemented based on their open-source repositories (see~\cref{appendix:baseline_details}). The high-confidence proxy datasets ($\mathcal{D}_{clean}$ and $\mathcal{D}_{poison}$) are extracted following~\cref{sec:aux_datasets}.
To train proxy models for selection, we use \textbf{IMDB}~\citep{zhang2015character} for the \sst victim model, \textbf{Twitter Abusive}~\citep{founta2018large} for the \textbf{OLID}, \textbf{SNLI}~\citep{young-etal-2014-image} for \textbf{MNLI}, and \textbf{BBCNews}~\citep{greene2006practical} for \textbf{AGNews}.

Our method uses hyperparameter $\alpha$ in~\cref{eqn:objective} to control the trade-off between backdoor defense and task utility. By default, $\alpha$ is set to 0.4 but can be adjusted based on priorities: smaller values (\eg 0.1) for lower ASR and larger values (\eg 1.0) for higher utility. The stopping patience $T=5$ is default in~\cref{alg:greedy_search}. In~\cref{appendix:sensitivity_hyperparameter}, we justify these choices and show that {\textit{our method is NOT particularly sensitive to hyperparameters}}.

\subsection{Main results}\label{sec:main_results}
\cref{tab:avg_roberta} presents the performance of our method compared to all baselines under benchmark backdoor attacks on the \textit{RoBERTa-large} model, with scores averaged over three runs using different seeds. Results on other datasets and architectures are provided in~\cref{appendix:diff_arch}. %We \colorbox{cellgood}{\textbf{highlight}} the top-2 ASR performances. 
Across all datasets and backdoor attacks, our method consistently ranks among the top 2, regarding backdoor removal performance, with minor harm to clean accuracy. In particular, \textbf{\textit{under the two particularly challenging backdoor tasks, LWS and HiddenKiller, \ours demonstrates significant improvements (i.e., at least 25\%) over all baselines}}--for example, for SST-2, our method reduces the ASR on LWS to 9.7\%, compared to 58.8\% for the following baseline Z-Def.
While Z-Def achieves competitive performance with \ours in defending against BadNet and InsertSent, Z-Def is much less effective against LWS on both datasets. We attribute this to Z-Def.'s reliance on lexical and syntactic features to detect outliers in the poisoned dataset, whereas LWS attacks subtly replace words with synonyms, effectively bypassing outlier detection. As for PURE, another competitive recent approach, we found its head-pruning step is highly unstable across both tasks and architectures. For example, it performs well on the BERT-base model in~\cref{tab:bert_result} (following the settings reported in \pure~\citep{zhao2024defense}), but performs poorly on most tasks with RoBERTa-large. \revision{Unlearning-based methods (\textbf{ABL}~\cite{li2021anti}) exhibit similar issues--often degrading CACC more than reducing ASR. Such instability has also been noted in prior works~\cite{liu2018fine, wu2022backdoorbench},  which shows that these methods are sensitive to both hyperparameter choices and the quality of the proxy data.} Additionally, we observed that all model-merging baselines suffer in defense performance when merging with a single proxy model, aligning with the results in~\citet{arora-etal-2024-heres}, and tend to prioritize preserving accuracy over removing backdoors. We discuss the potential reasons for this and the instability of PURE in~\cref{appendix:diff_arch}. \revision{We also compare our approach with two earlier model purification baselines, Fine-mixing~\cite{zhang2022fine} and Fine-purifying~\cite{zhang2023diffusion} in~\cref{appendix:more_baselines}.}
% For example, it performs well on BERT-base model in~\cref{tab:bert_result} (using the same structure and setting as \pure~\citep{zhao2024defense}), but overall poorly for RoBERTa-large. 
% This instability arises from PURE using accuracy on the clean proxy dataset as the stop criterion for pruning. For well-trained models with high accuracy, a substantial number of heads are pruned, leading to a broken purified model (\eg 0\% ASR but random-guess-level CACC).

% \input{assets/cross_domain}
% \input{assets/merge_dirty_complete}
\paragraph{Performance on LLMs.}  \label{sec:llm_result} 
\revision{Given the growing popularity of LLMs, we evaluate the performance of our method on Llama-2-7B, Mistral-7B and Qwen-2.5-7B for the SST-2 dataset (\cref{tab:llm_results}). Remarkably, in all cases, \ours effectively removes backdoors.}

%%%%%%%%%%%%%%%%%%%%%%%%%%%
\begin{table}[!hb]
% \vspace{-10pt}
\small
\centering
\renewcommand{\arraystretch}{1.1}
\setlength{\tabcolsep}{5.5pt}
\scalebox{0.9}{
\begin{tabular}{lcc}
\toprule
\textbf{Proxy Model Backdoor} & \textbf{CACC$\uparrow$} & \textbf{ASR$_{Victim}\downarrow$} \\
\midrule
Hidden Killer & 95.4 & \textbf{4.5} \\
BadNet (Diff. Trigger) & 87.8 & \textbf{6.9} \\
BadNet (Same Trigger) & 95.8 & 100.0 \\
\midrule
Victim Model (BadNet) & 95.6 & 100.0 \\
\bottomrule
\end{tabular}
}
\caption{Proxy models with implanted backdoors can still effectively purify victim models if the backdoor is \textit{not identical in both attack strategy and trigger}.}
\label{tab:proxy_backdoor_model}
\end{table}
\subsection{Ablation studies}\label{sec:ablation}
We next examine the key property of \ours--robustness across diverse proxy datasets. 
For completeness, we defer additional results to the appendix: 
\textit{stability} under different hyperparameter settings (\cref{appendix:sensitivity_hyperparameter}), 
\textit{transferability} across attacks (\cref{appendix:transfer}), 
preservation of clean victim model utility (\cref{appendix:clean_victim}), 
and resilience to re-tuning attacks (\cref{appendix:retuning}).

\begin{table*}[!t]
\small
\centering
\renewcommand{\arraystretch}{1.2}
\setlength{\tabcolsep}{5pt}
\scalebox{0.9}{
\begin{tabular}{cccccccccc}
\toprule
& & \multicolumn{2}{c}{\textbf{BadNet}} & \multicolumn{2}{c}{\textbf{InsertSent}} & \multicolumn{2}{c}{\textbf{LWS}} & \multicolumn{2}{c}{\textbf{HiddenKiller}} \\
\cmidrule(lr){3-4} \cmidrule(lr){5-6} \cmidrule(lr){7-8} \cmidrule(lr){9-10}
\multirow{-2}{*}{\textbf{Model}} & \multirow{-2}{*}{\textbf{Method}} & \textbf{ASR}$\downarrow$ & \textbf{CACC}$\uparrow$ & \textbf{ASR}$\downarrow$ & \textbf{CACC}$\uparrow$ & \textbf{ASR}$\downarrow$ & \textbf{CACC}$\uparrow$ & \textbf{ASR}$\downarrow$ & \textbf{CACC}$\uparrow$ \\
\midrule %%%%%%%%%%%%%%% LLAMA 2 SST2  %%%%%%%%%%%%%%%
& \cellcolor{cellbg}Benign & \cellcolor{cellbg}3.6 & \cellcolor{cellbg}96.7 & \cellcolor{cellbg}3.5 & \cellcolor{cellbg}96.7  & \cellcolor{cellbg}14.9 & \cellcolor{cellbg}96.7  & \cellcolor{cellbg}15.2 & \cellcolor{cellbg} 96.7\\
& \cellcolor{cellbg}Victim & \cellcolor{cellbg}100.0 & \cellcolor{cellbg}97.0 & \cellcolor{cellbg}100.0 & \cellcolor{cellbg}97.1 & \cellcolor{cellbg}98.7 & \cellcolor{cellbg}96.1 & \cellcolor{cellbg}95.9 & \cellcolor{cellbg}96.7 \\
% & \cellcolor{cellbg}Proxy Model (IMDB) & \cellcolor{cellbg}4.7 & \cellcolor{cellbg}90.8 & \cellcolor{cellbg}7.1 &  \cellcolor{cellbg}90.8 & \cellcolor{cellbg}15.5 &  \cellcolor{cellbg}90.8& \cellcolor{cellbg}6.4 & \cellcolor{cellbg}90.8 \\
\cline{2-10}
\multirow{-4}{*}{Llama-2} & \ours & \textbf{3.5} & 91.7 & \textbf{3.6} & 93.4 & \textbf{12.2} & 91.5 & \textbf{7.6} & 91.1 \\
\midrule %%%%%%%%%%%%%%% Mistral 2 SST2 %%%%%%%%%%%%%%%
& \cellcolor{cellbg}Benign &  \cellcolor{cellbg}3.8	&  \cellcolor{cellbg}97.1	& \cellcolor{cellbg}4.5		  & \cellcolor{cellbg}97.1  & \cellcolor{cellbg}7.7		 &  \cellcolor{cellbg}97.1 &  \cellcolor{cellbg}12.3	& \cellcolor{cellbg}97.1 \\
& \cellcolor{cellbg}Victim & \cellcolor{cellbg}100.0 & \cellcolor{cellbg}95.8 & \cellcolor{cellbg}100.0 & \cellcolor{cellbg}96.5 & \cellcolor{cellbg}98.9 & \cellcolor{cellbg}95.5 & \cellcolor{cellbg}95.9 & \cellcolor{cellbg}96.4 \\
% & \cellcolor{cellbg}Proxy Model (IMDB) & \cellcolor{cellbg}5.0 & \cellcolor{cellbg}91.1 & \cellcolor{cellbg}9.5 &  \cellcolor{cellbg}91.1 & \cellcolor{cellbg}5.5 & \cellcolor{cellbg}91.1  & \cellcolor{cellbg}6.1 & \cellcolor{cellbg}91.1 \\
\cline{2-10}
\multirow{-4}{*}{Mistral} & \ours & \textbf{4.4} & 92.1 & \textbf{7.1} & 92.4 & \textbf{3.8} & 91.9 & \textbf{5.3} & 91.9 \\
\midrule %%%%%%%%%%%%%%% Qwen 2.5-7B SST2  %%%%%%%%%%%%%%%
& \cellcolor{cellbg}Benign & \cellcolor{cellbg}5.0 & \cellcolor{cellbg}97.0 & \cellcolor{cellbg}2.7 & \cellcolor{cellbg}97.0  & \cellcolor{cellbg}15.1 & \cellcolor{cellbg}97.0  & \cellcolor{cellbg}15.7 & \cellcolor{cellbg} 97.0\\
& \cellcolor{cellbg}Victim & \cellcolor{cellbg}100.0 & \cellcolor{cellbg}96.3 & \cellcolor{cellbg}100.0 & \cellcolor{cellbg}92.9 & \cellcolor{cellbg}98.5 & \cellcolor{cellbg}98.5 & \cellcolor{cellbg}96.0 & \cellcolor{cellbg}96.2 \\
% & \cellcolor{cellbg}Proxy Model (IMDB) & \cellcolor{cellbg}4.7 & \cellcolor{cellbg}90.8 & \cellcolor{cellbg}7.1 &  \cellcolor{cellbg}90.8 & \cellcolor{cellbg}15.5 &  \cellcolor{cellbg}90.8& \cellcolor{cellbg}6.4 & \cellcolor{cellbg}90.8 \\
\cline{2-10}
\multirow{-4}{*}{Qwen-2.5} & \ours & \textbf{6.3} & 94.5 & \textbf{7.5} & 94.4 & \textbf{16.4} & 91.7 & \textbf{15.0} & 93.0 \\
\bottomrule
\end{tabular}
}
\caption{Performance of our method on \textit{Llama-2-7b}, \textit{Mistral-7b} and \textit{Qwen-2.5-7b} on \sst dataset, averaged over three seeds.}
\vspace{-4pt}
\label{tab:llm_results}
\end{table*}

\begin{table*}[!t]
\small
\centering
\renewcommand{\arraystretch}{1.2}
\scalebox{0.9}{
\begin{tabular}{cccccccccc}
\toprule
& & \multicolumn{2}{c}{\textbf{BadNet}} & \multicolumn{2}{c}{\textbf{InsertSent}} & \multicolumn{2}{c}{\textbf{LWS}} & \multicolumn{2}{c}{\textbf{HiddenKiller}} \\
\cmidrule(lr){3-4} \cmidrule(lr){5-6} \cmidrule(lr){7-8} \cmidrule(lr){9-10}
\multirow{-2}{*}{\textbf{Victim Model}} & \multirow{-2}{*}{\textbf{Proxy Model}} & \textbf{ASR} $\downarrow$ & \textbf{CACC} $\uparrow$ & \textbf{ASR} $\downarrow$ & \textbf{CACC} $\uparrow$ & \textbf{ASR} $\downarrow$ & \textbf{CACC} $\uparrow$ & \textbf{ASR} $\downarrow$ & \textbf{CACC} $\uparrow$ \\
\hline %%%%%%%%%%%%%%% SST-2  %%%%%%%%%%%%%%%
& IMDB & 5.4 & 92.8 & 1.6 & 94.0 & 12.6 & 93.8 & 11.0 & 93.0 \\
& YELP & 4.3 & 96.8 & 8.1 & 94.8 &  23.2 & 92.3 & 19.1 & 93.8 \\
\multirow{-3}{*}{SST-2} & AMAZON & 3.8 & 92.9 & 1.1 & 95.8 & 20.1 & 92.6  &  13.5 & 93.8 \\
\bottomrule
\end{tabular}
}
% \vspace{-8pt}
\caption{Proxy models trained on different datasets can all effectively purify victim models (trained on the SST-2).}
\vspace{-8pt}
\label{tab:cross_domain}
\end{table*}

\paragraph{Proxy-model robustness:} \ours retains the robustness property of model-merging defenses: it is resilient to both the specific choice of the proxy model and the benign or malicious nature of the proxy model; \ie as long as the backdoors in the two models are not identical, the backdoor can be effectively removed.
We examine the sensitivity of \ours to the selection of proxy models in two scenarios. \textbf{\textit{Scenario 1: Different proxies. }}We evaluated the performance of using different proxy models trained on homologous datasets, such as \textit{IMDB}, \textit{Yelp}, and \textit{Amazon}~\citep{zhang2015character}, to purify a victim model trained on \textit{SST-2}. As shown in~\cref{tab:cross_domain}, our method is largely insensitive to the choice of proxy model across all backdoor attacks. Using any proxy model, \ours effectively defends against all tested attacks.
\textbf{\textit{Scenario 2: Backdoored proxies. }} We tested cases where the victim model and the proxy model were trained on the same dataset (SST2) with different backdoor attacks (BadNet and Hidden Killer) or on different datasets (SST2 and IMDB) with the same attacks (BadNet). \move{From~\cref{tab:proxy_backdoor_model} and~\cref{tab:merge_dirty_complete}, we observed that \textbf{\textit{as long as the backdoor in proxy model is not identical to that in victim model (i.e., the same backdoor strategy with an identical backdoor trigger), \ours consistently mitigates the backdoor}}. We attribute this to the combination of module-level substitution and disruptive nature of model merging, which breaks the backdoor pathway such that a non-identical proxy cannot restore it.}

\begin{table}[!t]
\small
\centering
\renewcommand{\arraystretch}{1.1}
\setlength{\tabcolsep}{6pt}
\scalebox{0.9}{
\begin{tabular}{cc|cc}
\toprule
\multicolumn{2}{c|}{\textbf{Sub-proxy datasets}} & \multicolumn{2}{c}{\textbf{GMS}} \\
\cmidrule(lr){1-2} \cmidrule(lr){3-4}
\textbf{Proxy Clean} & \textbf{Proxy Poison} & \textbf{CACC} $\uparrow$ & \textbf{ASR} $\downarrow$ \\
\midrule
Random Sample & Outlier Detection & 93.6 & \textbf{3.2} \\
Random Sample & Oracle & 94.9 & \textbf{4.1} \\
Oracle & Outlier Detection & 92.7 & \textbf{4.3} \\
Oracle & Oracle & 95.6 & \textbf{4.3} \\
\midrule
\multicolumn{2}{c|}{Benign SST-2} & 95.6 & 4.1 \\
\bottomrule
\end{tabular}
}
\caption{Performance comparison of different sub-proxy dataset configurations on GMS.}
\vspace{-10pt}
\label{tab:proxy_dataset}
\end{table}

\paragraph{Proxy-data robustness:} Prior pruning- and unlearning-based methods strictly rely on access to clean data or known backdoor triggers~\cite{min2024uncovering, li2021anti, chen2022effective, li2023reconstructive}, typically extracted from the training subset. In contrast, \ours operates under more practical settings, \eg post-processing after suboptimal data detections, where the data received is a mixture of clean and poisoned samples. Our previous experiments used $\mathcal{D}_{clean}$ from random sampling (containing 20\% poisoned data) and $\mathcal{D}_{poison}$ from heuristic outlier detection. \cref{tab:proxy_clean_ratio} shows that even when 70\% of the ``clean'' proxy dataset is poisoned, \ours is still effective in purification. In~\cref{appendix:proof_random_sampling}, we mathematically justify this and discuss the analytical constraints on proxy datasets. These results demonstrate that our method is highly robust to proxy-data construction.
Nevertheless, \cref{tab:proxy_dataset} shows that more accurate proxy datasets--such as an oracle clean or label-flipped poison dataset--can further enhance performance, enabling precise module identification and full utility preservation (\eg CACC of 95.6\%). However, even without optimal proxies, our method consistently removes backdoors, achieving benign-level ASR (4.1\%) with comparable CACC.  

%%%%%%%%%%%%%%%%%%%%%%%%%%%%%%%%%%%%%%%%%%%%%%

\begin{table}[!t]
\small
\centering
\renewcommand{\arraystretch}{1.1}
\setlength{\tabcolsep}{6pt}
\scalebox{0.9}{
\begin{tabular}{c|cc|cc}
\toprule
 & \multicolumn{2}{c|}{\textbf{SST-2}} & \multicolumn{2}{c}{\textbf{AGNews}} \\
\cmidrule(lr){2-3} \cmidrule(lr){4-5}
 \multirow{-2}{*}{\textbf{Ratio $\rho$}} & \textbf{CACC} $\uparrow$ & \textbf{ASR} $\downarrow$ & \textbf{CACC} $\uparrow$ & \textbf{ASR} $\downarrow$ \\
\midrule
0.00 & 93.58 & 3.38 & 91.76 & 2.42 \\
% 0.10 & 93.58 & 3.38 & 91.76 & 2.42 \\
% 0.20 & 91.06 & 4.28 & 91.59 & 3.61 \\
0.30 & 91.51 & 3.15 & 91.59 & 3.61 \\
% 0.40 & 89.11 & 5.18 & 91.59 & 3.61 \\
% 0.50 & 89.11 & 5.18 & 91.59 & 3.61 \\
0.60 & 89.11 & 5.18 & 88.67 & 1.42 \\
0.70 & 89.11 & 5.18 & 88.67 & 1.42 \\
0.80 & 89.11 & 5.18 & 94.59 & \textbf{99.75} \\
0.90 & 89.11 & 5.18 & 94.83 & \textbf{99.82} \\
\bottomrule
\end{tabular}
}
\caption{Purification performance as the ratio $\rho$ of poisoned data in extracted $\mathcal{D}_{clean}$ increases.}
\vspace{-10pt}
\label{tab:proxy_clean_ratio}
\end{table}

\section{Conclusion} \raggedbottom
We propose \ourmethod, \revision{a model purification defense} against data-poisoning attacks in NLP \revision{based on guided merging of the victim model with a single (existing) proxy model.}
\revision{Extensive experiments on both standard NLP models and LLMs validate its effectiveness, particularly against challenging attacks like LWS and HiddenKiller. Beyond strong empirical results, we validated desirable properties of \ours: robustness to proxy selection, tolerance to imperfect data, stability across hyperparameters, and transferability across attacks. These findings suggest that guided single-proxy merging offers a practical alternative to retraining-based defenses.}
% The process leverages the semantic knowledge encoded in proxy models trained on homologous tasks. 

\section*{Limitations}\label{sec:limitations}
\revision{Our method advances model-merging-based defenses by guiding the merging process with a trade-off signal between task utility and backdoor risk, reducing the requirement from a number of proxy models to just one. 
One potential limitation is that, like prior proxy-model-based defenses, we focus on the homogeneous setting and do not explore heterogeneous proxies. We make this choice for two reasons. First, in the defense scenario, merging serves solely as a tool for purification rather than an end in itself, so defender can control which proxy model to use. In practice, obtaining a homogeneous proxy is not difficult, making the heterogeneous case less relevant for our scenarios. Second, our goal is to introduce and validate the concept of guided merging for backdoor defense, rather than to develop state-of-the-art merging techniques. Since knowledge transfer across heterogeneous models itself remains an open challenge~\citep{xu2024training,du2025adamms}, we leave such extensions to future work.
Moreover, while our method tolerates the use of a malicious proxy model, as long as its backdoor is not identical to the victim’s (\ie the same attack method with the exact same trigger, which is rare in practice), as shown in~\cref{tab:merge_dirty_complete}, \revision{further relaxing this, such as enabling defenses against identically backdoored proxies or developing proxy-free approaches, could lead to even more robust and powerful defense strategies.}}

\revision{
\section*{Ethical statement} \label{sec:ethical_statement}
Our method introduces a retraining-free purification approach for backdoored models, contributing to mitigating security risks in NLP systems from backdoor attacks. Additionally, our defense approach reduces the need for reannotating datasets to ensure they are purely clean, minimizing the potential harm of exposing annotators to harmfully poisoned contents. While we do not anticipate any direct negative consequences from this work, we hope it inspires further advancements in the development of robust, retraining-free defense methods for more realistic scenarios in future research.
}

\section*{Acknowledgements}
We acknowledge the support from 2024 FSE Strategic Startup and the credits awarded from Google Cloud Platform.
% \section*{Ethical considerations}
% \yao{optional}
% \bibliography{latex/custom,references}
\bibliography{references}
% \bibliography{custom}

\appendix
% \onecolumn
\newpage

\section{Experiment details}
\subsection{Experiment setup} We adopt the same training configuration as WAG~\citep{arora-etal-2024-heres} to ensure a fair comparison. This includes fine-tuning \textit{RoBERTa-large}, \textit{BERT-base-uncased}, \revision{as well as LLMs including \textit{Llama 2 7B}~\citep{touvron2023llama}, \textit{Mistral 7B}~\citep{jiang2023mistral}, and \textit{Qwen 2.5 7B}~\citep{qwen2.5}} on poisoned datasets, using the Adam optimizer with no weight decay~\citep{kingma2014adam} and a learning rate of $2 \times 10^{-5}$. For encoder models, the batch size, maximum sequence length, and epoch are set to 32, 128, and 3, respectively. \revision{For decoder LLMs, due to computational limits, we adapt low-rank adaptation (LoRA)~\citep{hu2021lora} for all the linear modules within the Transformer layers and apply our method to substitute those LoRA modules after training. We train these LLMs for 2 epochs with a batch size of 8 and a maximum sequence length of 128.}

To validate our defense, we first adapt the aforementioned backdoor methods and datasets to obtain victim models. \move{Specifically, to train backdoored models, we construct poisoned datasets by injecting triggers into the training split, while the validation split is used to construct a clean and a poison test set respectively. For MNLI, we use a random subset of 100,000 samples to reduce overhead.} Next, we train proxy models on homologous datasets to prepare for future merging. We then construct high-confidence proxy-clean ($\mathcal{D}_{clean}$) and proxy-poison ($\mathcal{D}_{poison}$) sets from the training data (using the methods illustrated in~\cref{sec:aux_datasets}) to guide our parameter substitution strategy search. 

For a model trained on unverified data, access to fully clean data is often not guaranteed. However, the corresponding task is typically known to the defender, providing an opportunity to identify homologous models trained on overlapping tasks that can be used for merging. For instance, previous works, WAG and PURE, utilized the \textbf{IMDB}~\citep{maas-etal-2011-learning} dataset to purify models trained on poisoned \textbf{SST-2} datasets, as both share the sentiment classification task domain.

In this work, we select proxy models that share a similar domain with the poisoned tasks. Specifically, we use \textbf{Twitter Abusive}~\citep{founta2018large} for \textbf{OLID}, \textbf{SNLI}\footnote{Creative Commons Attribution-ShareAlike 4.0 International License}~\citep{young-etal-2014-image} for \textbf{MNLI}, and \textbf{BBCNews}~\citep{greene2006practical} for \textbf{AGNews}. To evaluate generalizability, we use three sentiment classification datasets--\textbf{IMDB}, \textbf{YELP}, and \textbf{Amazon Reviews}~\citep{zhang2015character}--to purify victim \textbf{SST-2} models through model merging. All datasets are downloaded from Hugging Face, which adheres to the Apache License 2.0.

Our method involves only one hyperparameter, $\alpha$, as described in ~\cref{eqn:objective}, which controls the preference between model utility and attack resistance during the model merging strategy search. We set $\alpha$ to 0.4 by default, which slightly favors seeking a more attack-resistant model. It can be adjusted to smaller values (\eg 0.1) when a lower ASR is the primary target, and to larger values (\eg 1.0) when high utility is demanded. 

\subsection{Computational Resources}\label{appendix:compute_resources} We conduct experiments using three seeds on a single \revision{A100} GPU, and report the average scores. Running our method takes only 4 minutes on a single GPU for a 24-layer \textit{RoBERTa-large} architecture.
%For the \textit{roberta-large} model, the seeds are 42, 1000, and 2000, while for \textit{bert-base}, the seeds are 1000, 2000, and 3000. We run experiments with \textit{bert-base} and seed 42 for \textit{roberta-large} on a single Nvidia V100 GPU, while seeds 1000 and 2000 for \textit{roberta-large} are run on an A1000 GPU.
% All experiments are conducted on a single V100 GPU. 

\begin{table*}[!t]
\small
\centering
\renewcommand{\arraystretch}{1}
\setlength{\tabcolsep}{5.5pt}
\begin{tabular}{cccccccccc}
\toprule
& & \multicolumn{2}{c}{\textbf{BadNet}} & \multicolumn{2}{c}{\textbf{InsertSent}} & \multicolumn{2}{c}{\textbf{LWS}} & \multicolumn{2}{c}{\textbf{HiddenKiller}} \\
\cmidrule(lr){3-4} \cmidrule(lr){5-6} \cmidrule(lr){7-8} \cmidrule(lr){9-10}
\multirow{-2}{*}{\textbf{Dataset}} & \multirow{-2}{*}{\textbf{Method}} & \textbf{ASR}$\downarrow$ & \textbf{CACC}$\uparrow$ & \textbf{ASR}$\downarrow$ & \textbf{CACC}$\uparrow$ & \textbf{ASR}$\downarrow$ & \textbf{CACC}$\uparrow$ & \textbf{ASR}$\downarrow$ & \textbf{CACC}$\uparrow$ \\
\midrule %%%%%%%%%%%%%%% SST-2  %%%%%%%%%%%%%%%
& \cellcolor{cellbg}Benign & \cellcolor{cellbg}4.1 & \cellcolor{cellbg}95.9 & \cellcolor{cellbg}2.2 & \cellcolor{cellbg}95.9 & \cellcolor{cellbg}12.8 & \cellcolor{cellbg}95.9 & \cellcolor{cellbg}16.5 & \cellcolor{cellbg}95.9 \\
& \cellcolor{cellbg}Victim & \cellcolor{cellbg}100.0 & \cellcolor{cellbg}96.0 & \cellcolor{cellbg}100.0 & \cellcolor{cellbg}96.3 & \cellcolor{cellbg}98.0 & \cellcolor{cellbg}95.4 & \cellcolor{cellbg}96.5 & \cellcolor{cellbg}95.7 \\
& \cellcolor{cellbg}Proxy Model (IMDB) & \cellcolor{cellbg}7.4 & \cellcolor{cellbg}89.1 & \cellcolor{cellbg}4.3 & \cellcolor{cellbg}89.1 & \cellcolor{cellbg}10.8 & \cellcolor{cellbg}89.1 & \cellcolor{cellbg}13.7 & \cellcolor{cellbg}89.1 \\
\cline{2-10}
& ONION & 56.8 & 92.9 & 99.9 & 93.3 & 85.7 & 91.9 & 92.9 & 92.9 \\
& Z-Def. & \cellcolor{cellgood}\textbf{4.6} & \textbf{96.1} & \cellcolor{cellgood}\textbf{1.8} & 95.6 & 97.3 & 95.3 & \cellcolor{cellgood}\textbf{35.7} & 95.4 \\
& PURE & 0 & 50.9 & 0 & 50.9 & 0 & 50.9 & 0 & 50.9 \\
\cline{2-10}
& TIES & 99.9 & 95.7 & 100.0 & 95.8 & 93.5 & 95.2 & 88.8 & 95.3 \\
& DARE w/ TIES  & 99.3 & 96.0 & 100.0 & \textbf{96.2} & 96.4 & \textbf{95.7} & 92.7 & \textbf{95.8} \\
& WAG & 84.4 & 94.8 & 60.1 & 95.2  & \cellcolor{cellgood}\textbf{58.8} & 94.8 & 56.2 & 92.5 \\
\multirow{-8}{*}{SST-2} & Ours (\ours) & \cellcolor{cellgood}\textbf{4.5} & 91.6 & \cellcolor{cellgood}\textbf{1.9} & 92.5 & \cellcolor{cellgood}\textbf{9.7} & 91.7 & \cellcolor{cellgood}\textbf{10.4} & 91.2 \\
\midrule %%%%%%%%%%%%%%% OLID  %%%%%%%%%%%%%%%
& \cellcolor{cellbg}Benign & \cellcolor{cellbg}29.0 & \cellcolor{cellbg}84.9 & \cellcolor{cellbg}31.0 & \cellcolor{cellbg}84.9 & \cellcolor{cellbg}45.6 & \cellcolor{cellbg}84.9 & \cellcolor{cellbg}53.6 & \cellcolor{cellbg}84.9 \\
& \cellcolor{cellbg}Victim & \cellcolor{cellbg}99.9 & \cellcolor{cellbg}85.2 & \cellcolor{cellbg}100.0 & \cellcolor{cellbg}85.0 & \cellcolor{cellbg}94.5 & \cellcolor{cellbg}85.1 & \cellcolor{cellbg}100.0 & \cellcolor{cellbg}85.0 \\
& \cellcolor{cellbg}Proxy Model (Twitter) & \cellcolor{cellbg}37.8 & \cellcolor{cellbg}84.4 & \cellcolor{cellbg}40.0 & \cellcolor{cellbg}84.4 & \cellcolor{cellbg}55.4 & \cellcolor{cellbg}84.4 & \cellcolor{cellbg}67.1 & \cellcolor{cellbg}84.4 \\
\cline{2-10}
& ONION & 75.0 & 84.8 & 99.4 & 84.8 & 86.1 & 84.1 & 99.6 & 84.7 \\
& Z-Def. & \cellcolor{cellgood}\textbf{29.7} & 85.6 & \cellcolor{cellgood}\textbf{30.3} & 85.3 & 93.5 & 85.3 & \cellcolor{cellgood}\textbf{53.9} & 85.7 \\
& PURE & 82.2 & 75.2 & 87.5 & 75.2 & 76.1 & 79.1 & 100.0 & 72.1 \\
\cline{2-10}
& TIES & 36.2 & 84.8 & 38.2 & 84.8 & \cellcolor{cellgood}\textbf{57.5} & 84.5 & 65.3 & 84.5 \\
& DARE w/ TIES  & 95.6 & \textbf{86.3} & 77.8 & \textbf{86.0} & 90.5 & \textbf{85.9} & 86.8 & \textbf{86.3} \\
& WAG & 52.0 & 85.0 & 48.6 & 84.6 & 63.1 & 84.5 & 68.2 & 85.0 \\
\multirow{-8}{*}{OLID} & Ours (\ours)$^{*}$ & \cellcolor{cellgood}\textbf{32.1} & 85.0 & \cellcolor{cellgood}\textbf{28.0} & 84.2 & \cellcolor{cellgood}\textbf{52.2} & 84.4 & \cellcolor{cellgood}\textbf{64.2}  & 84.9 \\
\midrule  %%%%%%%%%%%%%%% MNLI  %%%%%%%%%%%%%%%
& \cellcolor{cellbg}Benign & \cellcolor{cellbg}12.3 & \cellcolor{cellbg}87.6 & \cellcolor{cellbg}12.6 & \cellcolor{cellbg}87.6 & \cellcolor{cellbg}26.4 & \cellcolor{cellbg}87.6 & \cellcolor{cellbg}36.9 & \cellcolor{cellbg}87.6 \\
& \cellcolor{cellbg}Victim & \cellcolor{cellbg}100.0 & \cellcolor{cellbg}89.4 & \cellcolor{cellbg}100.0 & \cellcolor{cellbg}90.3 & \cellcolor{cellbg}96.0 & \cellcolor{cellbg}89.0 & \cellcolor{cellbg}99.9 & \cellcolor{cellbg}89.4 \\
& \cellcolor{cellbg}Proxy Model (SNLI) & \cellcolor{cellbg}12.2 & \cellcolor{cellbg}84.1 & \cellcolor{cellbg}9.2 & \cellcolor{cellbg}84.1 & \cellcolor{cellbg}25.3 & \cellcolor{cellbg}84.1 & \cellcolor{cellbg}31.7 & \cellcolor{cellbg}84.1 \\
\cline{2-10}
& ONION & 64.3 & 86.1 & 98.6 & 86.9 & 89.0 & 85.5 & 98.8 & 86.6 \\
& Z-Def. & \cellcolor{cellgood}\textbf{11.1} & 88.3 & \cellcolor{cellgood}\textbf{11.6} & 89.7 & 92.2 & 89.1 & \cellcolor{cellgood}\textbf{50.6} & 89.7 \\
& PURE & 33.3 & 33.8 & 33.3 & 33.8 &33.3 & 33.8 & 33.3 & 33.8 \\
\cline{2-10}
& TIES & 91.6 & 89.2 & 94.5 & 89.5 & 80.7 & 89.2 & 88.6 & 90.0 \\
& DARE w/ TIES & 93.5 & \textbf{90.5} & 100.0 & \textbf{91.4} & 93.9 & \textbf{89.9} & 99.4 & \textbf{90.5} \\
& WAG & 71.0 & 88.7 & 60.3 & 88.5 & \cellcolor{cellgood}\textbf{77.9} & 88.0 & 80.5 & 88.8 \\
\multirow{-8}{*}{MNLI} & Ours (\ours) & \cellcolor{cellgood}\textbf{10.8} & 86.5 & \cellcolor{cellgood}\textbf{10.7} & 86.3 & \cellcolor{cellgood}\textbf{14.0} & 86.5 & \cellcolor{cellgood}\textbf{31.7} & 86.3 \\
\midrule %%%%%%%%%%%%%%% AGNews %%%%%%%%%%%%%%%
& \cellcolor{cellbg}Benign & \cellcolor{cellbg}1.9 & \cellcolor{cellbg}95.4 & \cellcolor{cellbg}0.5 & \cellcolor{cellbg}95.4 & \cellcolor{cellbg}0.5 & \cellcolor{cellbg}95.4 & \cellcolor{cellbg}1.1 & \cellcolor{cellbg}95.4 \\
& \cellcolor{cellbg}Victim & \cellcolor{cellbg}99.9 & \cellcolor{cellbg}95.1 & \cellcolor{cellbg}99.6 & \cellcolor{cellbg}95.3 & \cellcolor{cellbg}99.6 & \cellcolor{cellbg}94.5 & \cellcolor{cellbg}100.0 & \cellcolor{cellbg}95.1 \\
& \cellcolor{cellbg}Proxy Model (BBCNews) & \cellcolor{cellbg}1.5 & \cellcolor{cellbg}70.2 & \cellcolor{cellbg}1.7 & \cellcolor{cellbg}70.2 & \cellcolor{cellbg}1.8 & \cellcolor{cellbg}70.2 & \cellcolor{cellbg}3.4 & \cellcolor{cellbg}70.2 \\
\cline{2-10}
& ONION & 59.4 & 94.8 & 97.8 & 95.1 & 84.8 & 94.5 & 99.6 & 94.7 \\
& Z-Def. & \cellcolor{cellgood}\textbf{1.6} & \textbf{95.3} & \cellcolor{cellgood}\textbf{0.4}& 95.3 & 97.9 & \textbf{96.1} & 100.0 & 95.0 \\
& PURE & 2.8 & 86.3 & 3.0 & 85.4 & \cellcolor{cellgood}\textbf{6.5} & 85.0 & \cellcolor{cellgood}\textbf{9.4} & 84.6 \\
\cline{2-10}
& TIES & 99.9 & 94.6 & 99.6 & 94.4 & 97.7 & 95.8 & 100.0 & 94.4 \\
& DARE w/ TIES & 99.9 &  95.2& 99.6 & \textbf{95.4} & 97.8 & 96.5  & 100.0 & \textbf{95.2} \\
& WAG & 92.7 & 94.1 & 97.8 & 94.4 & 78.0 & 93.9 & 90.9 & 94.3 \\
\multirow{-8}{*}{AGNews} & Ours (\ours) & \cellcolor{cellgood}\textbf{2.5} & 91.0 & \cellcolor{cellgood}\textbf{2.4} & 92.6 & \cellcolor{cellgood}\textbf{3.2} & 91.7 & \cellcolor{cellgood}\textbf{6.5} & 90.4 \\
\bottomrule
\end{tabular}
\caption{Performance of our method compared to baselines under various backdoor attacks on the RoBERTa-large model, with each value averaged over three seeds.}
\label{tab:avg_roberta_all}
\end{table*}
% \weijun{Settings: \textcolor{red}{roberta-large}, Poison Rate 20\%, Training for 3 Epochs, merging hyper-param alpha is 0.4, seed is 42/1000/2000, ave. score.}

\subsection{Implementation details for baselines}\label{appendix:baseline_details} We follow the open-source implementations for each baseline, and basically using their default hyper-parameters, while maintaining using the identical datasets, backdoor settings and the trained models for a fair comparison.

For the baseline \textbf{DARE}~\cite{yu2024language}, it first applies random parameter dropping and rescaling to the involved models with a specified drop rate, and then incorporates model merging techniques to combine the processed models. Various model merging methods can be integrated with DARE, and we choose TIES merging as a representative, as it demonstrates decent performance for encoder-based models (\eg \textit{bert-base, roberta-base}). Their method involves three tunable hyperparameters for encoder-based LMs, as outlined in Table 5 of their paper: \textit{drop rate}, \textit{scaling term}, and \textit{ratio to retain}. We retained the original search space for \textit{drop rate} and \textit{scaling term} but expanded the \textit{ratio to retain} from [0.1, 0.2, 0.3] to [0.1, 0.2, 0.3, 0.4, 0.5, 0.6, 0.7, 0.8] for the \textit{roberta-large} model to ensure more complete search.

For the baseline \textbf{PURE}~\citep{zhao2024defense}, it first trains the victim model on a proxy \textit{clean} dataset (from the same domain or a similar task, \eg poisoned on \textbf{IMDB} and fine-tuned on \sst) and prunes the attention heads that cause attention drifting between poisoned and clean texts. The pruned model is then fine-tuned further on the proxy clean dataset to normalize the remaining attention heads for purification. To ensure a fair comparison in the main experiments with the \textit{roberta-large} model, we use the same poisoned and proxy datasets, \eg poisoned on the SST-2 dataset with IMDB as the proxy clean dataset. We set the hyperparameter accuracy threshold (used to stop head pruning) to 0.90 for \sst, \mnli, and \agnews, and 0.85 for \olid to prevent overly aggressive pruning. For the \textit{bert-base} model, we follow the original implementation, including the use of the \sst dataset for fine-tuning SST-2 victim models and maintaining the default accuracy threshold of 0.85. While PURE uses the \textit{label flip rate} (\textbf{LFR}) as its evaluation metric for backdoor defense (implanting triggers into test data while keeping the labels unchanged), we adopt the attack success rate (\textbf{ASR}) on label-flipped test data as our evaluation metric for a fair comparison with our method.

%To ensure a fair comparison, we use the same poisoned and proxy datasets as in our experiments, \eg poisoned on the SST-2 dataset and using IMDB as the proxy clean dataset. For the hyperparameter accuracy threshold used to stop head pruning, we follow the default setting of 0.85 in their implementation for \textit{bert-base}. However, for \textit{roberta-large} on the \sst, \mnli, and \agnews datasets, we adjust the threshold to 0.90 to prevent overly aggressive pruning, as \textit{roberta-large} exhibits higher initial accuracy scores. For the \olid dataset, we keep the threshold at 0.85. While PURE uses the \textit{label flip rate} (\textbf{LFR}) as its evaluation metric for backdoor defense (implanting triggers into test data while keeping the labels unchanged), we adopt the attack success rate (\textbf{ASR}) on label-flipped test data as our evaluation metric for a fair comparison with our method.

We adhere to the default implementations and hyperparameter settings for all other baseline methods.

% \yao{@Weijun: Could we add the details of the hyperparameters for the baselines here, especially if any of them were implemented differently from their original settings?}
% \weijun{Yes, for PURE, we use proxy datasets, such as IMDB, rather than SST-2, for a fair comparison. The default acc threshold for pure is 0.85, but we set it to 0.9 for SST-2, MNLI and AGNEWS, for roberta-large to avoid too aggressive pruning.}

\section{Additional results}

\begin{table*}[!t]
\small
\centering
\renewcommand{\arraystretch}{1.1}
\setlength{\tabcolsep}{5.5pt}
\begin{tabular}{cccccccccc}
\toprule
& & \multicolumn{2}{c}{\textbf{BadNet}} & \multicolumn{2}{c}{\textbf{InsertSent}} & \multicolumn{2}{c}{\textbf{LWS}} & \multicolumn{2}{c}{\textbf{HiddenKiller}} \\
\cmidrule(lr){3-4} \cmidrule(lr){5-6} \cmidrule(lr){7-8} \cmidrule(lr){9-10}
\multirow{-2}{*}{\textbf{Dataset}} & \multirow{-2}{*}{\textbf{Method}} & \textbf{ASR}$\downarrow$ & \textbf{CACC}$\uparrow$ & \textbf{ASR}$\downarrow$ & \textbf{CACC}$\uparrow$ & \textbf{ASR}$\downarrow$ & \textbf{CACC}$\uparrow$ & \textbf{ASR}$\downarrow$ & \textbf{CACC}$\uparrow$ \\
\midrule
& \cellcolor{cellbg}Benign & \cellcolor{cellbg}8.5 & \cellcolor{cellbg}92.9 & \cellcolor{cellbg}3.5 & \cellcolor{cellbg}92.9 & \cellcolor{cellbg}22.2 & \cellcolor{cellbg}92.9 & \cellcolor{cellbg}17.3 & \cellcolor{cellbg}92.9 \\
& \cellcolor{cellbg}Victim & \cellcolor{cellbg}100.0 & \cellcolor{cellbg}92.9 & \cellcolor{cellbg}100.0 & \cellcolor{cellbg}92.2 & \cellcolor{cellbg}98.1 & \cellcolor{cellbg}91.5 & \cellcolor{cellbg}95.9 & \cellcolor{cellbg}91.8 \\
& \cellcolor{cellbg}Proxy Model (IMDB) & \cellcolor{cellbg}10.4 & \cellcolor{cellbg}85.4 & \cellcolor{cellbg}6.4 & \cellcolor{cellbg}85.4 & \cellcolor{cellbg}17.9 & \cellcolor{cellbg}85.4 & \cellcolor{cellbg}14.7 & \cellcolor{cellbg}85.4 \\
\cline{2-10}
& ONION & 58.0 & 89.9 & 99.7 & 89.9 & 85.4 & 88.6 & 94.4 & 89.2 \\
& Z-Def. & \cellcolor{cellgood}\textbf{8.3} & 92.5 & \cellcolor{cellgood}\textbf{1.8} & 91.9 & 97.4 & 90.9 & 38.7 & 91.5 \\
% & PURE~\citep{zhao2024defense} & 36.3 & 91.6 & \cellcolor{cellgood}\textbf{7.9} & 91.7 & \cellcolor{cellgood}\textbf{73.4} & 91.3 & \cellcolor{cellgood}\textbf{29.3} & 91.3 \\
& PURE & 30.1 & 92.0 & 10.7 & 91.5 & \cellcolor{cellgood}\textbf{66.4} & 91.1 & \cellcolor{cellgood}\textbf{28.7} & 91.3 \\
\cline{2-10}
% & DAN & \textbf{0.0} & 92.1 & \textbf{0.0} & 91.3 & 17.2 & 90.6 & 49.6 & 90.9 \\
& TIES & 99.2 & 92.7 & 98.2 & \textbf{92.4} & 91.2 & 92.3 & 86.8 & \textbf{92.6} \\
& DARE w/ TIES & 100.0 & \textbf{93.0} & 90.8 & 88.9 & 92.6 & \textbf{92.7} & 94.1 & 90.6 \\
& WAG & 74.3 & 91.9 & 70.7 & 91.7 & 73.5 & 91.7 & 60.3 & 91.6 \\
% \multirow{-9}{*}{SST-2} & Ours & 10.8 (11.3) & 86.8 (87.1) & 9.8 (10.7) & 88.8 (89.3) & 27.4 (27.4) & 88.0 (88.0) & 19.5 (22.6) & 86.8 (88.2) \\
\cline{2-10}
\multirow{-9}{*}{SST-2} & Ours (\ours) & \cellcolor{cellgood}\textbf{10.8} & 86.8 & \cellcolor{cellgood}\textbf{9.8}  & 88.8 & \cellcolor{cellgood}\textbf{27.4} & 88.0 & \cellcolor{cellgood}\textbf{19.5} & 86.8 \\
\bottomrule
\end{tabular}
\caption{Performance of our method compared to baselines under various backdoor attacks on the BERT-base
model, with each value averaged over three seeds. The victim model is trained on SST-2 dataset.}
\label{tab:bert_result}
\end{table*}
\subsection{Performance across different model architectures and different dataset}\label{appendix:diff_arch}
\begin{figure*}[ht]
    \centering
    \includegraphics[width=\linewidth]{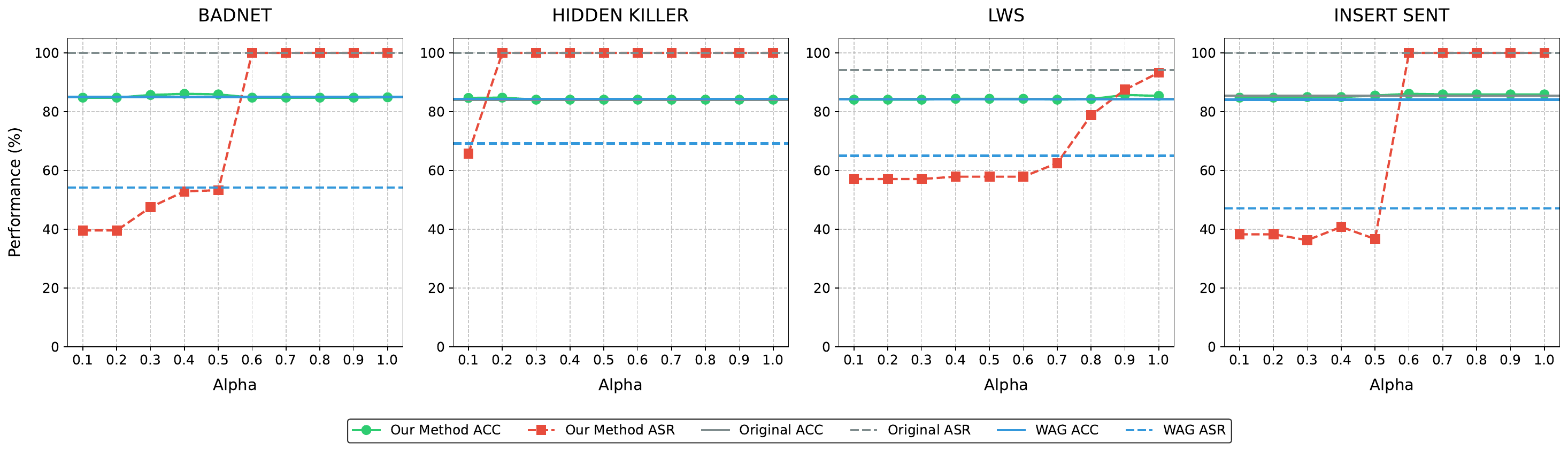}
    \caption{Results of using different weights (\textit{alpha}) on OLID dataset.}
    \label{fig:alpha_grid_olid}
% \end{figure*}
% \begin{figure*}
    \centering
    \includegraphics[width=\linewidth]{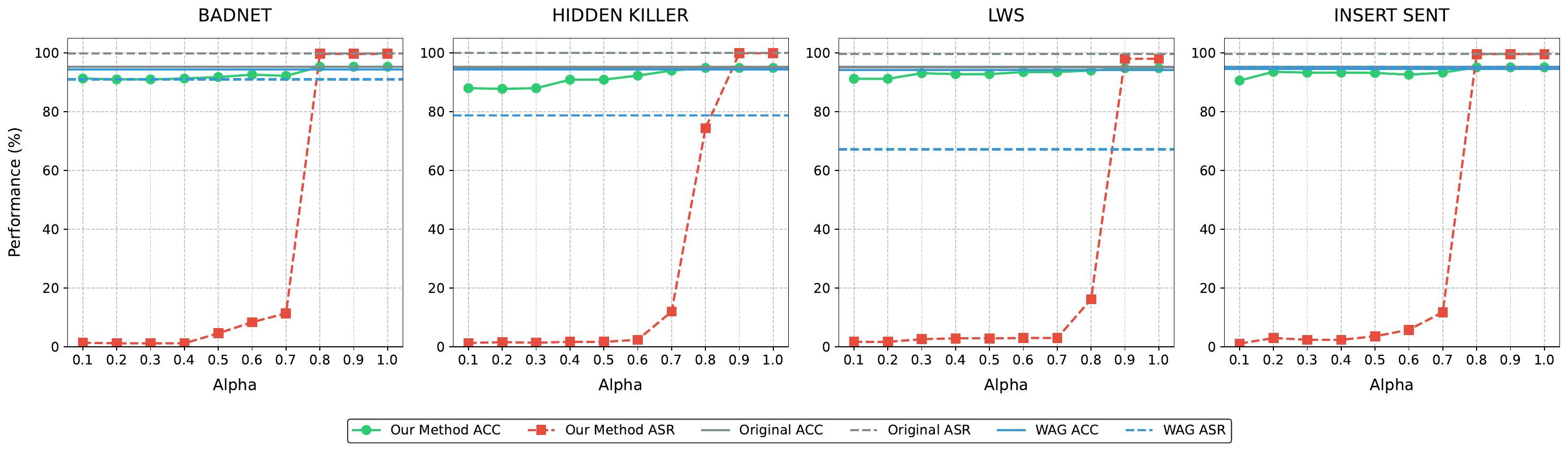}
    \caption{Results of using different weights (\textit{alpha}) on AGNEWS dataset.}
    \label{fig:alpha_grid_agnews}
\end{figure*}

\noindent \textbf{Performance across different dataset.} Due to space constraints, we present the complete performance results of our method compared to all baselines under various backdoor attacks on the \textit{RoBERTa-large} model across all four datasets in~\cref{tab:avg_roberta}. The settings are consistent with those in~\cref{sec:main_results}, with each score averaged over three runs using different seeds. For the OLID dataset, we specifically set \(\alpha=0.1\) to enable more aggressive defense, as shown in~\cref{fig:alpha_grid_olid}. The top two ASR performances are highlighted. Across all datasets and backdoor attacks, our method consistently ranks among the top two in backdoor removal performance, with minimal harm to clean accuracy. Notably, under the two particularly challenging backdoor tasks, LWS and HiddenKiller, \ours achieves significant improvements (at least 25\%) over all baselines. For instance, on SST-2, \ours reduces the ASR for LWS to 9.7\%, compared to 58.8\% for the next-best baseline, Z-Def.

\move{While Z-Def achieves competitive performance with \ours in defending against BadNet and InsertSent, Z-Def is much less effective against LWS on both datasets. This is because Z-Def relies on lexical and syntactic features to detect outliers in the poisoned dataset, whereas LWS attacks subtly replace words with synonyms, effectively bypassing outlier detection. As for PURE, another competitive recent approach, we found its head-pruning step is highly unstable across both tasks and architectures. For example, it performs well on the BERT-base model in~\cref{tab:bert_result} (following the settings reported in \pure~\citep{zhao2024defense}), but performs poorly on most tasks with RoBERTa-large.
This instability seems to arise from PURE's reliance on accuracy from the clean proxy dataset as the pruning stopping criterion. For well-trained models with high accuracy, a substantial number of heads are pruned, leading to a broken purified model (\eg 0\% ASR but random-guess-level CACC).}

\move{Additionally, although merging multiple models has been shown to be an effective backdoor defense, we observed that applying all merging baselines using a single proxy model is sub-optimal, aligning with the results in~\citet{arora-etal-2024-heres}. We found these methods prioritize preserving accuracy over removing backdoors. This behavior likely arises from the design of merging mechanisms, which are intended to preserve the performance of each merged model on downstream tasks. Consequently, backdoor tasks are treated equivalently to downstream tasks, exposing a persistent vulnerability.}

\vspace{10pt}
\textbf{Performance across different model architectures.} We also evaluated our method on different architectures. \cref{tab:bert_result} shows the results on the BERT-base model. The poison rate remains at 20\%, and each model is trained for three epochs. Scores are reported under \(\alpha=0.4\) and averaged over three seeds: 1000, 2000, and 3000. Across all attacks, our method performs consistently well, particularly excelling against more challenging attack strategies. Notably, it achieves over a 39\% advantage in defending against LWS and a 9\% improvement over the second-best baseline for HiddenKiller.

% \revision{\subsection{Comparisons to Fine-mixing~\cite{zhang2022fine} and Fine-purifying~\cite{zhang2023diffusion}} 

\revision{\subsection{Comparisons to other baseline defenses}  \label{appendix:more_baselines}
\noindent \textbf{Model purification baselines}
Fine-mixing~\cite{zhang2022fine} and Fine-purification~\cite{zhang2023diffusion} are two additional model purification baselines. We compare our method against their reported performance in their papers\footnote{The code is not publicly available yet.}, strictly following their experimental settings. Specifically, we evaluate under the same conditions using two types of attacks (BadNet and InsertSent) and two model architectures (BERT-base and RoBERTa-base) on the AGNews dataset. As shown in~\cref{tab:fine_mixing_purifying}, our method consistently achieves better defense performance across all settings. Notably, against the more advanced InsertSent attack, our approach reduces the attack success rate (ASR) by at least 17\% in absolute values compared to Fine-mixing and Fine-purification.\begin{table*}[!t]
\small
\centering
\renewcommand{\arraystretch}{1.1}
\setlength{\tabcolsep}{5.5pt}
\begin{tabular}{cccc|ccc}
\toprule
\textbf{Architecture} & \textbf{Methods} & \multicolumn{2}{c}{\textbf{BadNet}} & \multicolumn{2}{c}{\textbf{InsertSent}} \\
\cmidrule(lr){3-4} \cmidrule(lr){5-6}
& & \textbf{CACC} $\uparrow$ & \textbf{ASR} $\downarrow$ & \textbf{CACC} $\uparrow$ & \textbf{ASR} $\downarrow$ \\
\midrule
\multirow{3}{*}{BERT-base} 
& Fine-mixing$^{*}$ & 90.17 & 12.32 & 90.40 & 32.37 \\
& Fine-purifying$^{*}$ & 90.86 & 3.3 & 91.13 & 23.69 \\
& Ours & \textbf{91.8} & \textbf{1.8} & \textbf{86.49} & \textbf{7.00} \\
\midrule
\multirow{3}{*}{RoBERTa-base} 
& Fine-mixing$^{*}$ & 86.39 & 18.12 & 86.11 & 35.97 \\
& Fine-purifying$^{*}$ & 86.64 & 17.56 & 86.85 & 19.20 \\
& Ours & \textbf{91.2} & \textbf{1.63} & \textbf{89.62} & \textbf{1.42} \\
\bottomrule
\end{tabular}
\caption{Performance comparison of our method against Fine-mixing~\cite{zhang2022fine} and Fine-purifying~\cite{zhang2023diffusion} on AGNews dataset. The best results are in \textbf{bold}. Results in rows marked with $^{*}$ are taken directly from the respective papers.}
\label{tab:fine_mixing_purifying}
\end{table*}
}

\revision{\paragraph{Poison sample detection with retraining: SEEP~\cite{he2024seep}}
While this paper focuses on the important but often overlooked post-deployment setting, it is natural to ask: \textit{\textbf{if we had access to a highly effective data detection method capable of identifying most poisoned samples, how would our method compare?}}
To explore this, we evaluate against SEEP~\cite{he2024seep}, a state-of-the-art poison detection method, on defending against BadNet attacks on the SST-2 dataset. As shown in~\cref{tab:seep_comparison}, although SEEP performs well under this relatively simple attack, our method--despite relying only on a naive heuristic for proxy data selection--achieves even better performance.
As further discussed in~\cref{tab:proxy_dataset}, incorporating more advanced prior knowledge (\eg by using stronger detection methods to construct sub-proxy datasets) can further enhance the effectiveness of our approach.
\begin{table}[!t]
\small
\centering
\renewcommand{\arraystretch}{1.1}
\setlength{\tabcolsep}{5.5pt}
\begin{tabular}{llcc}
\toprule
\textbf{Architecture} & \textbf{Defense Method} & \textbf{CACC}$\uparrow$ & \textbf{ASR}$\downarrow$ \\
\midrule
RoBERTa-large & SEEP with retraining & 95.5 & \textit{24.3} \\
              & \textbf{Ours} & 91.7 & \textbf{9.7} \\
\midrule
BERT-base     & SEEP with retraining & 92.4 & 29.4 \\
              & \textbf{Ours} & 88.0 & \textbf{27.4} \\
\bottomrule
\end{tabular}
\caption{Comparison of our method with retraining-based defenses on BERT-base and RoBERTa-large for SST-2 dataset.}
\label{tab:seep_comparison}
\end{table}
}

\revision{\subsection{Comparisons of all defenses under a more recent attack: BITE~\cite{yan20223bite}}\label{appendix:bite}
BITE~\cite{yan20223bite} is a recent insertion-based textual backdoor attack that leverages label-biased tokens as stealthy triggers. \cref{tab:bite_defense} shows the performance of our method and four baselines in defending against BITE on the SST-2 dataset. Our method consistently ranks among the top two in defense effectiveness, achieving comparable or even lower ASR than the benign model, while maintaining clean accuracy.}

\revision{We exclude BITE from the main results in~\cref{sec:main_results} due to a peculiar issue: BITE tends to yield an unusually high ASR even on benign models. This undermines the reliability of ASR as an evaluation metric in this setting and makes it difficult to confidently interpret defense performance. We attribute this to BITE’s trigger selection process, which biases trigger tokens toward naturally label-correlated words, effectively exploiting model priors without requiring explicit poisoning.
\begin{table*}[!t]
\small
\centering
\renewcommand{\arraystretch}{1.1}
\setlength{\tabcolsep}{5.5pt}
\begin{tabular}{llcccc}
\toprule
\multirow{2}{*}{\textbf{Model}} & \multirow{2}{*}{\textbf{Defense Method}} & \multicolumn{2}{c}{\textbf{BERT-base}} & \multicolumn{2}{c}{\textbf{RoBERTa-Large}} \\
\cmidrule(lr){3-4} \cmidrule(lr){5-6}
& & \textbf{CACC}$\uparrow$ & \textbf{ASR}$\downarrow$ & \textbf{CACC}$\uparrow$ & \textbf{ASR}$\downarrow$ \\
\midrule
\multirow{2}{*}{Victim (BITE)} 
& Benign Model & 92.9 & 41.4 & 95.9 & 38.3 \\
& No Defense & 92.6 & 80.7 & 95.6 & 81.0 \\
\midrule
\multirow{5}{*}{Defense}
& Z-Defense & 92.3 & 51.9 & 95.2 & 44.9 \\
& ONION & 89.7 & 70.5 & 92.9 & 68.2 \\
& ABL & 92.0 & 82.1 & \textbf{51.6} & \cellcolor{cellgood}\textbf{1.0} \\
& WAG & 92.9 & \cellcolor{cellgood}\textbf{47.5} & 94.4 & 45.2 \\
& \textbf{Ours} & 88.9 & \cellcolor{cellgood}\textbf{49.9} & 90.3 & \cellcolor{cellgood}\textbf{27.9} \\
\bottomrule
\end{tabular}
\caption{Defense performance against the BITE~\cite{yan20223bite} attack on BERT-base and RoBERTa-large models for SST-2 dataset.}
\label{tab:bite_defense}
\end{table*}}

\subsection{Performance under different poison rate: 20\%, 10\% and 5\%} \label{appendix:diff_poison_ratio}
While the experiments in the main paper are conducted with a 20\% poison rate to ensure fair comparison with the baselines, following the settings in~\citep{Dai2019ABA,Qi2021HiddenKI,arora-etal-2024-heres}, \ie 20\% of the training samples are poisoned, we also evaluate lower poison rates of 10\% and 5\%, as shown in~\cref{tab:poison_rate}. Our method is effective in model purification across all poison rates.
% \begin{table*}[ht]
% \centering
% \begin{tabular}{ccccccccc}
% \toprule
% \multirow{2}{*}{Dataset} & \multirow{2}{*}{Backdoor} & \multirow{2}{*}{Proxy Model} & \multicolumn{2}{c}{20\%} & \multicolumn{2}{c}{10\%} & \multicolumn{2}{c}{5\%} \\
% \cmidrule(lr){4-5} \cmidrule(lr){6-7} \cmidrule(lr){8-9}
%  &  &  & \textbf{ASR}$\downarrow$ & \textbf{CACC}$\uparrow$ & \textbf{ASR}$\downarrow$ & \textbf{CACC}$\uparrow$ & \textbf{ASR}$\downarrow$ & \textbf{CACC}$\uparrow$ \\
% \midrule
% SST-2 & BadNet & IMDB & 4.5 & 91.6 & 4.5 & 94.4 & 12.4 & 91.9 \\
% \bottomrule
% \end{tabular}
% \caption{Performance results with varying poison rates.}
% \label{tab:poison_rate}
% \end{table*}

\begin{table}[!b]
\small
\centering
\begin{tabular}{ccccc}
\toprule
\textbf{Victim} & \textbf{Proxy} & \textbf{Poison Rate} & \textbf{Metric} & \textbf{Result} \\
\midrule
 & \multirow{6}{*}{IMDB} 
& \multirow{2}{*}{20\%} & \textbf{ASR}$\downarrow$ & 4.5 \\
 &    &  & \textbf{CACC}$\uparrow$ & 91.6 \\
\cline{3-5}
 SST-2 &    & \multirow{2}{*}{10\%} & \textbf{ASR}$\downarrow$ & 4.5 \\
(BadNet) &    &  & \textbf{CACC}$\uparrow$ & 94.4 \\
\cline{3-5}
 &    & \multirow{2}{*}{5\%}  & \textbf{ASR}$\downarrow$ & 12.4 \\
 &    &  & \textbf{CACC}$\uparrow$ & 91.9 \\
\bottomrule
\end{tabular}
\caption{Results for SST-2 with varying poison rates.}
\label{tab:poison_rate}
\end{table}

% \subsection{Does the choice of proxy models matter?} Our method relies on a proxy model for substitution, raising the natural question of \textit{how sensitive it is to the selection of proxy models}. To address this, we tested two scenarios: \textit{\textbf{Scenario 1: }}We evaluated the performance of using different proxy models trained on homologous datasets, such as \textit{IMDB}, \textit{Yelp}, and \textit{Amazon}~\citep{zhang2015character}, to purify a victim model trained on \textit{SST-2}. As shown in~\cref{tab:cross_domain}, our method is largely insensitive to the choice of proxy model across all backdoor attacks. Using any proxy model, \ours effectively defends against all tested attacks.
% \textit{\textbf{Scenario 2:}} What if the proxy model itself is backdoored? We tested cases where the victim model and the proxy model were trained on the same dataset with different backdoor attacks or on different datasets with the same attacks. From~\cref{tab:merge_dirty_complete}, we observed that as long as the backdoor in the proxy model is not identical to that in the victim model (\ie the same backdoor strategy with the same backdoor trigger), our \ours method consistently removes the backdoor.

\subsection{Stability: How sensitive is our method to the selection of hyperparameters?} \label{appendix:sensitivity_hyperparameter}
\begin{figure}
    \centering
    \includegraphics[width=\linewidth]{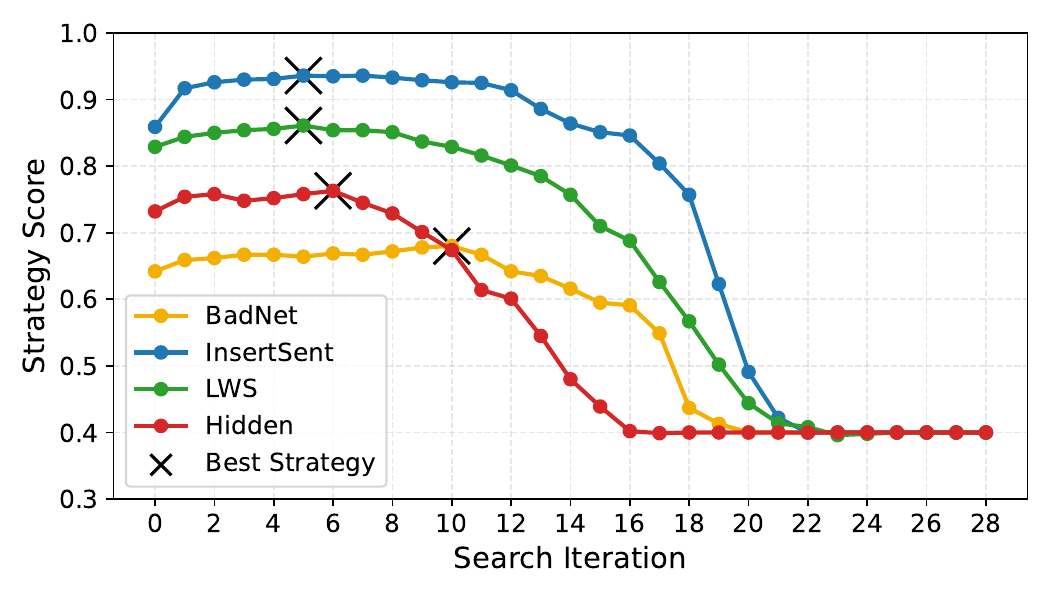}
    \caption{The search iteration history for four kinds of backdoors on \sst dataset.}
    \label{fig:strategy_score}
\end{figure}

We next show that \ours exhibits stable performance across a wide range of hyperparameters, in contrast to pruning-based and unlearning-based methods, which have been shown to be more sensitive to hyperparameter choices~\cite{liu2018fine, wu2022backdoorbench} (as also observed in~\cref{tab:avg_roberta}). Our method involves two hyperparameters: \(\alpha\) in~\cref{eqn:objective} and stopping patience \(T\), with default values of \(\alpha=0.4\) and \(T=5\) used in main results. Below, we justify these choices. As shown in~\cref{fig:alpha_grid_olid} and~\cref{fig:alpha_grid_agnews}, for most dataset and backdoor attack combinations, the default value \(\alpha = 0.4\) strikes a balanced trade-off between utility and backdoor defense strength. An exception is the HiddenKiller attack on the OLID dataset, where a more aggressive \(\alpha=0.1\) is recommended. This is because OLID is a dataset collected from tweets, while HiddenKiller paraphrases the data using formal syntactic templates, drastically altering OLID's language style. This change significantly impacts CACC, necessitating a higher weight for backdoor removal to ensure effective purification. For stopping patience, as shown in~\cref{fig:strategy_score}, the score computed in~\cref{eqn:objective} generally follows a (weakly) monotonically decreasing trend. While a larger \(T\) improves performance, we choose \(T=5\) as a reasonable balance between performance and efficiency.

% \begin{figure*}
%     \centering
%     \includegraphics[width=\linewidth]{figures/alpha_grid_olid.pdf}
%     \caption{Results of using different \textit{alpha} values on OLID dataset.}
%     \label{fig:alpha_grid_olid}
% % \end{figure*}
% % \begin{figure*}
%     \centering
%     \includegraphics[width=\linewidth]{figures/alpha_grid_agnews.pdf}
%     \caption{Results of using different \textit{alpha} values on AGNEWS dataset.}
%     \label{fig:alpha_grid_agnews}
% \end{figure*}

\begin{table}[!b]
\vspace{-10pt}
\small
\centering
\renewcommand{\arraystretch}{1.2}
\begin{tabular}{cccc}
\toprule
\textbf{Dataset} & \textbf{Model} & \textbf{CACC} $\uparrow$ \\
\midrule
\multirow{3}{*}{SST-2} 
& Benign Victim Model & 95.9 \\
& Proxy Model (IMDB) & 88.7 \\
\cline{2-3}
& Purified Model (\ours) & 95.8 \\
\midrule
\multirow{3}{*}{AGNews} 
& Benign Victim Model & 95.2 \\
& Proxy Model (BBCNews) & 78.1 \\
\cline{2-3}
& Purified Model (\ours) & 95.3 \\
\bottomrule
\end{tabular}
\caption{CACC of benign victim models preserved.} 
\label{tab:merge_clean}
\end{table}
% \textcolor{red}{roberta-large}. \textit{Alpha} is set to 0.4.}

% \begin{table*}[t]
% \small
% \centering
% \renewcommand{\arraystretch}{1.2}
% \begin{tabular}{ccccccc}
% \toprule
% \multirow{2}{*}{\textbf{Dataset}} & \multirow{2}{*}{\textbf{Model}} & \multirow{2}{*}{\textbf{CACC}$\uparrow$} & \multicolumn{4}{c}{\textbf{ASR}$\downarrow$} \\
% \cline{4-7}
% & & & BadNet & InsertSent & LWS & HiddenKiller \\
% \midrule
% \multirow{3}{*}{SST-2} 
% & Benign Model & 95.9 & 0.4 & 0.5 & 12.4 & 14.6 \\
% & Proxy Model (IMDB) & 88.7 & 6.5 & 7.0 & 10.6 & 10.8 \\
% \cline{2-7}
% & Merged Model & 95.8 & 3.2 & 0.7 & 11.7 & 12.2 \\
% \midrule
% \multirow{3}{*}{AGNews} 
% & Benign Model & 95.2 & 0.5 & 0.5 & 1.1 & 5.2 \\
% & Proxy Model (BBCNews) & 78.1 & 1.5 & 1.5 & 2.0 & 1.9 \\
% \cline{2-7}
% & Merged Model & 95.3 & 0.4 & 0.4 & 0.9 & 3.8 \\
% \bottomrule
% \end{tabular}
% \caption{Performance of merging two \textbf{benign} models on SST-2 and AGNEWS datasets using \textcolor{red}{roberta-large}. \textit{Alpha} is set to 0.4.}
% \label{tab:merge_clean}
% \end{table*}
\begin{table}[!b]
\small
\centering
\begin{tabular}{ccc}
\toprule
\textbf{Proxy Dataset} & \textbf{CACC}$\uparrow$ & \textbf{ASR}$\downarrow$ \\
\midrule
Pure clean proxy dataset & 88.3 & \textbf{4.7} \\
Mixed poisoned proxy dataset & 91.6 & \textbf{4.5} \\
\bottomrule
\end{tabular}
\caption{Performance of our method with a clean vs. mixed proxy dataset, for the \textit{roberta-large} model trained on the SST-2 dataset.}
\label{tab:proxy_overall_dataset}
\end{table}

\begin{table*}[!ht]
\centering
% \small
\begin{tabular}{lcccccc}
\toprule
 &  & \multicolumn{4}{c}{\textbf{Methods}} \\ 
\cmidrule(lr){3-6}
\multirow{-2}{*}{\textbf{Metrics}}   &   \multirow{-2}{*}{\textbf{Stage}}     & \textbf{WAG} & \textbf{PURE} & \textbf{Clean} & \textbf{Ours} \\ \midrule
& Before  & 94.72   & 93.12   & 95.76     & 91.86 \\
\multirow{-2}{*}{\textbf{CACC}$\uparrow$}  & After & 94.61  & 91.85  & 94.95           & 93.80 \\ \midrule
 & Before  & 84.40   & \ \ 6.31  & \ \ 4.05 & \ \ 3.83 \\
\multirow{-2}{*}{\textbf{ASR}$\downarrow$}    & After  & 98.19    & 19.37     & \ \ 9.50  & 11.93 \\ \bottomrule
\end{tabular}
\caption{Evaluating the robustness of different parameter purification methods towards the Retuning Attack (RA) using 0.1\% of the original poisoned training set (which contains 20\% poisoned data) for fine-tuning. We report the performance of each methods before and after the RA.}
\vspace{-5pt}
\label{tab:retuning}
\end{table*}

\begin{figure*}
    \centering
    \includegraphics[width=\linewidth]{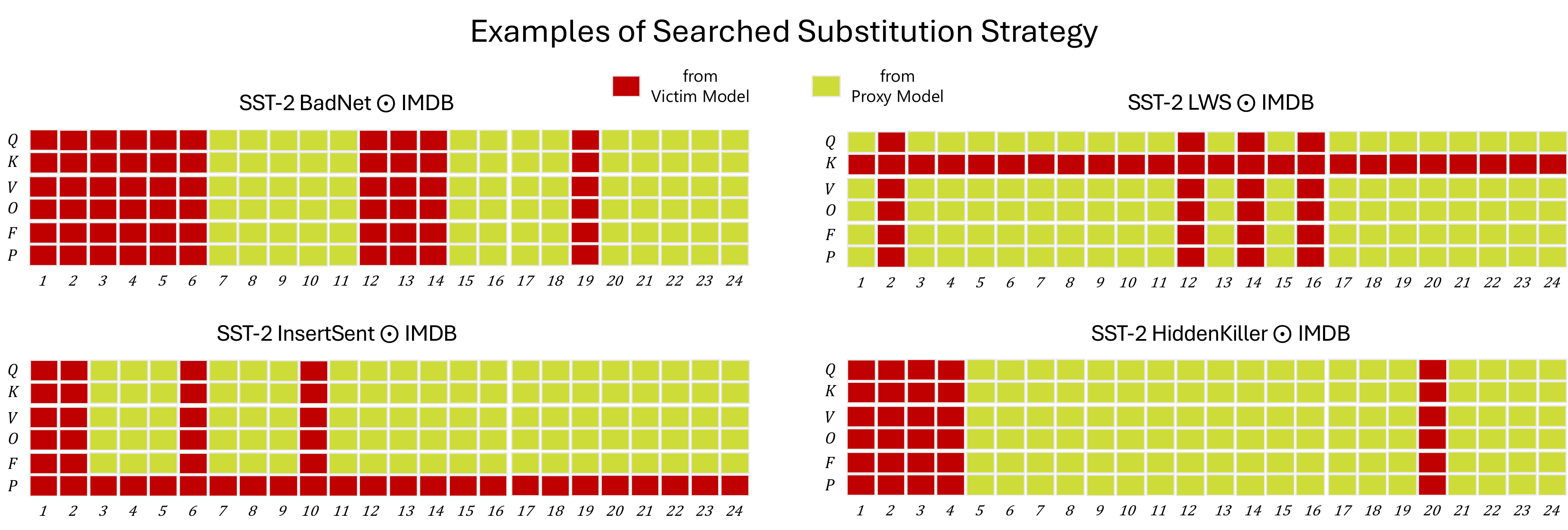}
    \caption{The optimal substitution strategy for defending against each backdoor attack for the \textit{roberta-large} model trained on the SST-2 dataset. The \textcolor{yellowgreen}{green} squares indicate the substituted modules.}
    \label{fig:transfer}
\end{figure*}

\move{\subsection{If the victim model is benign, will our method affect its utility?} \label{appendix:clean_victim}
While all our previous experiments focus on malicious victim models, it is also important to consider scenarios where the victim model is benign. \textit{Does our method compromise the utility of a clean victim model?} As shown in \cref{tab:merge_clean}, our results effectively dispel this concern. When applying \ours to a benign model, the CACC remains unaffected.
}

\revision{\subsection{Can our method be effective with a fully clean proxy dataset?} \label{appendix:clean_overall_dataset}
While we primarily consider the practical scenario of receiving a mixed dataset, \textit{is our method still effective if the received dataset is entirely clean?} 
As shown in \cref{tab:proxy_overall_dataset}, when provided only with a clean proxy dataset---without any information about the poisoned dataset---\ours can still effectively remove backdoors. This is because our substitution strategy starts by fully removing backdoors and then gradually recovering task utility.
}

\subsection{\ours remains effective as long as the proxy model is not backdoored with exactly the same attack and trigger.}\label{appendix:backdoor_proxy_model}
\begin{table*}[!t]
\small
\setlength{\tabcolsep}{8pt}
\centering
\begin{tabular}{ccccccc}
\toprule
\multirow{2}{*}{\textbf{Model}} & \multirow{2}{*}{\textbf{Dataset}} & \multirow{2}{*}{\textbf{Backdoored Method}} & \multicolumn{2}{c}{\textbf{Before Substitution}} & \multicolumn{2}{c}{\textbf{After Substitution}} \\
\cmidrule(lr){4-5} \cmidrule(lr){6-7}
& & & \textbf{CACC$\uparrow$} & \textbf{ASR$_{Victim}\downarrow$} & \textbf{CACC$\uparrow$} & \textbf{ASR$_{Victim}\downarrow$} \\
\midrule
Victim & SST-2 & BadNet & 95.6 & 100.0 & \multirow{2}{*}{95.3} & \multirow{2}{*}{4.5} \\
Proxy & SST-2 & HiddenKiller & 95.4 & 7.9 & &  \\
\midrule
Victim & SST-2 & HiddenKiller & 95.4 & 96.7 & \multirow{2}{*}{95.9} & \multirow{2}{*}{14.0} \\
Proxy & SST-2 & BadNet & 95.6 & 17.1 & &  \\
\midrule
Victim & SST-2 & \multirow{2}{*}{BadNet (Different triggers)} & 95.6 & 100.0 & \multirow{2}{*}{87.8} & \multirow{2}{*}{6.8} \\
Proxy & IMDB &  & 87.7 & 8.6 & &  \\
\midrule
Victim & SST-2 & \multirow{2}{*}{BadNet (Same triggers)} & 95.6 & 100.0 & \multirow{2}{*}{95.8} & \multirow{2}{*}{100.0} \\
Proxy & IMDB &  & 90.1 & 100.0 & &  \\
\bottomrule
\end{tabular}
\caption{Proxy models with implanted backdoors can still effectively purify victim models if the backdoor does not exactly match in both attack strategy and trigger.}
\vspace{-8pt}
\label{tab:merge_dirty_complete}
\end{table*}
An important question to consider is: What happens if the proxy model itself contains a backdoor? To investigate this, we considered scenarios where the victim and proxy models were either trained on the same dataset but with different backdoor attacks or on different datasets with the same attack.

From~\cref{tab:merge_dirty_complete}, our findings indicate that \ours remains effective in backdoor removal as long as the proxy model’s backdoor is not an exact match to that of the victim model, \ie both the same attack method and the same trigger.

\revision{
\subsection{Purification effectiveness as the poison ratio increases.}\label{appendix:proxy_clean_dataset}
Due to space constraints, we included an incomplete version of~\cref{tab:proxy_clean_ratio} in the main paper. Here, the full ratio table~\cref{tab:proxy_clean_ratio_full} provides a clearer trend: (1) As the proportion of poisoned data in the proxy ``clean'' dataset increases, the CACC of the purified model slightly decreases. This suggests that under a default weight of \(\alpha=0.4\) for \(\mathcal{D}_{clean}\), a higher level of poisoning may introduce confusion, leading \ours to mistakenly remove some clean task-critical modules due to their reduced relative weight (see~\cref{appendix:proof_random_sampling} for the analytical explanation). (2) However, even with poisoned ratios as high as 90\% in SST-2 and 70\% in AGNews, \ours remains effective in eliminating backdoor-related components and purifying the model. This effectiveness is mathematically justified in~\cref{appendix:proof_random_sampling}.
}

\begin{table}[!t]
\small
\centering
\renewcommand{\arraystretch}{1.1}
\setlength{\tabcolsep}{6pt}
\begin{tabular}{c|cc|cc}
\toprule
 & \multicolumn{2}{c|}{\textbf{SST-2}} & \multicolumn{2}{c}{\textbf{AGNews}} \\
\cmidrule(lr){2-3} \cmidrule(lr){4-5}
  \multirow{-2}{*}{\textbf{Ratio $\rho$}}  & \textbf{CACC} $\uparrow$ & \textbf{ASR} $\downarrow$ & \textbf{CACC} $\uparrow$ & \textbf{ASR} $\downarrow$ \\
\midrule
0.00 & 93.58 & 3.38 & 91.76 & 2.42 \\
0.10 & 93.58 & 3.38 & 91.76 & 2.42 \\
0.20 & 91.06 & 4.28 & 91.59 & 3.61 \\
0.30 & 91.51 & 3.15 & 91.59 & 3.61 \\
0.40 & 89.11 & 5.18 & 91.59 & 3.61 \\
0.50 & 89.11 & 5.18 & 91.59 & 3.61 \\
0.60 & 89.11 & 5.18 & 88.67 & 1.42 \\
0.70 & 89.11 & 5.18 & 88.67 & 1.42 \\
0.80 & 89.11 & 5.18 & 94.59 & \textbf{99.75} \\
0.90 & 89.11 & 5.18 & 94.83 & \textbf{99.82} \\
\bottomrule
\end{tabular}
\caption{Purification performance as the ratio $\rho$ of poisoned data in the proxy ``clean'' dataset $\mathcal{D}_{clean}$ increases.}
\label{tab:proxy_clean_ratio_full}
\end{table}

\subsection{Robustness to Retuning Attacks}\label{appendix:retuning}
% \yao{retuning ratio 0.2\%, roberta-large, IMDB relace SST2 badnet}
Recent works~\cite{min2024uncovering, qi2023fine} have highlighted a critical pitfall in backdoor defenses: purified models with low ASR often fail to completely eliminate inserted backdoor features. A straightforward Retuning Attack (RA)--which involves fine-tuning a purified model on a small number of backdoored samples (\eg 0.2\% of the original poisoned data) mixed with benign samples (to maintain clean Accuracy on the main task)--can easily recover the backdoor in the purified model, as demonstrated in~\cref{tab:retuning}. 

Several related studies~\cite{qi2023fine, chen2023janus, tarun2023fast} have sought to uncover the underlying reasons for this failure. A key observation is that the purified model’s parameters do not sufficiently deviate from those of the backdoored model. Specifically, the weight differences between a purified model and a (retrained) benign model are significantly larger than those between the purified and backdoored models. This suggests that purified and backdoored models are connected via a backdoor-related path in the loss landscape~\cite{min2024uncovering}. Such vulnerabilities pose severe risks, as purified models are often deployed in various downstream applications, and even with backdoor defenses applied, attackers can easily re-trigger the backdoor in downstream tasks~\cite{min2024uncovering}.

This challenge motivates us to propose a purification method aimed at breaking this backdoor-connected path. Intuitively, achieving this requires completely replacing "suspected" parameters inherited from the backdoored model, rather than pruning~\citep{zhao2024defense} or merging them with other parameters~\cite{arora-etal-2024-heres,yadav2024ties}. Moreover, because parameters often interact within functional modules, modifying at the module level rather than the individual parameter level tends to remove backdoor features more effectively. As shown in~\cref{tab:retuning}, when using 0.1\% of the original poisoned training set to launch the Retuning Attack (RA)~\cite{min2024uncovering}, the parameter-merging baseline, WAG~\cite{arora-etal-2024-heres}, is easily re-triggered. In contrast, our module-substitution method demonstrates greater robustness while maintaining utility, performing comparably to the clean model baseline (\ie finetuning a clean model on the retuning dataset). We provide our baseline PURE with additional advantages by granting access to the full clean dataset of the victim task for finetuning, as obtaining a usable purified model otherwise proves challenging. While the recent method PURE also demonstrates competitive performance, it is worth noting that significant effort was required to find a configuration where its pruning step did not break the purified model. Even with these adjustments, our method still substantially outperforms PURE, achieving an ASR of 11.93\% compared to 19.37\%--nearly twice the robustness.

\subsection{Examples of Searched Strategy}\label{appendix:examples}
We applied our \ours defense to purify victim models compromised by various backdoor attacks across four datasets. Examples of the searched strategies from the \sst dataset are shown in Figure~\ref{fig:transfer}. While the obtained strategies vary, we speculate that they are transferable, as adapting the minimum replacement strategy identified for BadNet to the other three attacks yields effective defense results, as discussed in~\cref{appendix:transfer}.

\begin{table*}[!t]
\small
\centering
\renewcommand{\arraystretch}{1.2}
\setlength{\tabcolsep}{5pt}
\begin{tabular}{cccccccccc}
\toprule
& & \multicolumn{2}{c}{\textbf{BadNet}} & \multicolumn{2}{c}{\textbf{InsertSent}} & \multicolumn{2}{c}{\textbf{LWS}} & \multicolumn{2}{c}{\textbf{HiddenKiller}} \\
\cmidrule(lr){3-4} \cmidrule(lr){5-6} \cmidrule(lr){7-8} \cmidrule(lr){9-10}
\multirow{-2}{*}{\textbf{Dataset}} & \multirow{-2}{*}{\textbf{Method}} & \textbf{ASR}$\downarrow$ & \textbf{CACC}$\uparrow$ & \textbf{ASR}$\downarrow$ & \textbf{CACC}$\uparrow$ & \textbf{ASR}$\downarrow$ & \textbf{CACC}$\uparrow$ & \textbf{ASR}$\downarrow$ & \textbf{CACC}$\uparrow$ \\
\midrule %%%%%%%%%%%%%%% SST-2  %%%%%%%%%%%%%%%
& \cellcolor{cellbg}Benign & \cellcolor{cellbg}4.1 & \cellcolor{cellbg}95.9 & \cellcolor{cellbg}2.2 & \cellcolor{cellbg}95.9 & \cellcolor{cellbg}12.8 & \cellcolor{cellbg}95.9 & \cellcolor{cellbg}16.5 & \cellcolor{cellbg}95.9 \\
& \cellcolor{cellbg}Victim & \cellcolor{cellbg}100.0 & \cellcolor{cellbg}96.0 & \cellcolor{cellbg}100.0 & \cellcolor{cellbg}96.3 & \cellcolor{cellbg}98.0 & \cellcolor{cellbg}95.4 & \cellcolor{cellbg}96.5 & \cellcolor{cellbg}95.7 \\
& \cellcolor{cellbg}Proxy Model (IMDB) & \cellcolor{cellbg}7.4 & \cellcolor{cellbg}89.1 & \cellcolor{cellbg}4.3 & \cellcolor{cellbg}89.1 & \cellcolor{cellbg}10.8 & \cellcolor{cellbg}89.1 & \cellcolor{cellbg}13.7 & \cellcolor{cellbg}89.1 \\
\cline{2-10}
 & \ours & 4.5 & 91.6 & 1.9 & 92.5 & 9.7 & 91.7 & 10.4 & 91.2 \\
\cline{2-10}
\multirow{-5}{*}{SST-2} & SST-2 BadNet Strategy &  4.5 & 91.6 &  9.8 & 93.0 &  22.3 & 92.8 &  13.2 & 90.3 \\
\midrule %%%%%%%%%%%%%%% OLID  %%%%%%%%%%%%%%%
& \cellcolor{cellbg}Benign & \cellcolor{cellbg}29.0 & \cellcolor{cellbg}84.9 & \cellcolor{cellbg}31.0 & \cellcolor{cellbg}84.9 & \cellcolor{cellbg}45.6 & \cellcolor{cellbg}84.9 & \cellcolor{cellbg}53.6 & \cellcolor{cellbg}84.9 \\
& \cellcolor{cellbg}Victim & \cellcolor{cellbg}99.9 & \cellcolor{cellbg}85.2 & \cellcolor{cellbg}100.0 & \cellcolor{cellbg}85.0 & \cellcolor{cellbg}94.5 & \cellcolor{cellbg}85.1 & \cellcolor{cellbg}100.0 & \cellcolor{cellbg}85.0 \\
& \cellcolor{cellbg}Proxy Model (Twitter) & \cellcolor{cellbg}37.8 & \cellcolor{cellbg}84.4 & \cellcolor{cellbg}40.0 & \cellcolor{cellbg}84.4 & \cellcolor{cellbg}55.4 & \cellcolor{cellbg}84.4 & \cellcolor{cellbg}67.1 & \cellcolor{cellbg}84.4 \\
\cline{2-10}
 & \ours & 32.1 & 85.0 & 28.0 & 84.2 & 52.2 & 84.4 & 64.2  & 84.9 \\
 \cline{2-10}
 \multirow{-5}{*}{OLID} & OLID BadNet Strategy & 33.6 & 85.0 &  37.4 & 84.6 &  56.5 & 84.3 &  65.0 & 84.7 \\
\midrule  %%%%%%%%%%%%%%% MNLI  %%%%%%%%%%%%%%%
& \cellcolor{cellbg}Benign & \cellcolor{cellbg}12.3 & \cellcolor{cellbg}87.6 & \cellcolor{cellbg}12.6 & \cellcolor{cellbg}87.6 & \cellcolor{cellbg}26.4 & \cellcolor{cellbg}87.6 & \cellcolor{cellbg}36.9 & \cellcolor{cellbg}87.6 \\
& \cellcolor{cellbg}Victim & \cellcolor{cellbg}100.0 & \cellcolor{cellbg}89.4 & \cellcolor{cellbg}100.0 & \cellcolor{cellbg}90.3 & \cellcolor{cellbg}96.0 & \cellcolor{cellbg}89.0 & \cellcolor{cellbg}99.9 & \cellcolor{cellbg}89.4 \\
& \cellcolor{cellbg}Proxy Model (SNLI) & \cellcolor{cellbg}12.2 & \cellcolor{cellbg}84.1 & \cellcolor{cellbg}9.2 & \cellcolor{cellbg}84.1 & \cellcolor{cellbg}25.3 & \cellcolor{cellbg}84.1 & \cellcolor{cellbg}31.7 & \cellcolor{cellbg}84.1 \\
\cline{2-10}
& \ours & 10.8 & 86.5 & 10.7 & 86.3 & 14.0 & 86.5 & 31.7 & 86.3 \\
\cline{2-10}
 \multirow{-5}{*}{MNLI} & MNLI BadNet Strategy & 12.2 & 86.3 &  10.8 & 86.3 &  27.0 & 85.2 &  35.2 & 86.4\\
\midrule %%%%%%%%%%%%%%% AGNews %%%%%%%%%%%%%%%
& \cellcolor{cellbg}Benign & \cellcolor{cellbg}1.9 & \cellcolor{cellbg}95.4 & \cellcolor{cellbg}0.5 & \cellcolor{cellbg}95.4 & \cellcolor{cellbg}0.5 & \cellcolor{cellbg}95.4 & \cellcolor{cellbg}1.1 & \cellcolor{cellbg}95.4 \\
& \cellcolor{cellbg}Victim & \cellcolor{cellbg}99.9 & \cellcolor{cellbg}95.1 & \cellcolor{cellbg}99.6 & \cellcolor{cellbg}95.3 & \cellcolor{cellbg}99.6 & \cellcolor{cellbg}94.5 & \cellcolor{cellbg}100.0 & \cellcolor{cellbg}95.1 \\
& \cellcolor{cellbg}Proxy Model (BBCNews) & \cellcolor{cellbg}1.5 & \cellcolor{cellbg}70.2 & \cellcolor{cellbg}1.7 & \cellcolor{cellbg}70.2 & \cellcolor{cellbg}1.8 & \cellcolor{cellbg}70.2 & \cellcolor{cellbg}3.4 & \cellcolor{cellbg}70.2 \\
\cline{2-10}
& \ours & 2.5 & 91.0 & 2.4 & 92.6 & 3.2 & 91.7 & 6.5 & 90.4 \\
\cline{2-10}
 \multirow{-5}{*}{AGNews} & AGNews BadNet Strategy & 1.4& 90.7 & 1.2 & 89.8 &  2.6 & 90.1 &  12.0 & 90.2\\
\bottomrule
\end{tabular}
\caption{The performance of transferring strategy searched based on BadNet to other attacks.}
\vspace{-4pt}
\label{tab:transfer_attack_complete}
\end{table*}

\subsection{Transferability of substitution strategies}  \label{appendix:transfer}
% We present the results for SST-2 and AGNews in~\cref{tab:transfer_attack}, with additional results for other datasets provided in~\cref{tab:transfer_attack_complete}. All experiments follow a consistent setup: the substitution strategy is optimized based on BadNet attacks and then applied to coordinate module substitutions for other attacks. For instance, in the MNLI task, the substitution strategy identified for defending against BadNet---using a proxy model trained on SNLI---is directly applied to replace modules in InsertSent, LWS, and HiddenKiller with those from the proxy model. Across all tasks, the defense performance is surprisingly strong, revealing an intriguing observation: a substantial overlap exists in the components impacted by different attacks.
Interestingly, we find that \ours exhibits strong transferability: a substitution strategy optimized for one attack often generalizes to others, enabling defense even without access to data. 
Visually, the module substitution strategies (\(S_{best}\)) identified by \ours across different backdoor attacks (examples in~\cref{fig:transfer} in~\cref{appendix:examples}), though distinct, share many similar patterns across attacks.

\move{
In~\cref{tab:transfer_attack_complete}, for each task dataset, we apply the substitution strategy found on BadNet (replacing the fewest modules and layers as shown in~\cref{fig:transfer}) to to determine which modules to substitute in the three other victim models under different attacks, replacing them with modules from their corresponding proxy models, \ie directly executing Line~\ref{alg_line:return} in~\cref{alg:greedy_search} using $S_{badnet}$. Compared with the complete \ours (\ie searching for the optimal strategies $S_{best}$ for each attack), the transferred strategy consistently performs comparably. This demonstrates that for defenders who only have access to the victim model and are unaware of the victim dataset, \textit{a universal \ours defense strategy exists for each task that can effectively defend against multiple attacks}.
}
\revision{\section{Constraints on the proxy datasets \(\mathcal{D}_{clean}\) and \(\mathcal{D}_{poison}\)} \label{appendix:proof_random_sampling}}

\revision{\cref{tab:proxy_clean_ratio} and~\cref{tab:proxy_dataset} illustrates that our method is quite robust to the construction of the proxy datasets \(\mathcal{D}_{clean}\) and \(\mathcal{D}_{poison}\), \eg even random sampling (for \(\mathcal{D}_{clean}\)) and inaccurate heuristics (for \(\mathcal{D}_{poison}\)) can yield effective defense. Natural questions arise: \textbf{Are there any constraints on the proxy datasets? Could the random sampling strategy for constructing \(\mathcal{D}_{clean}\) fail if \(\mathcal{D}_{poison}\) is highly inaccurate?} We next establish the constraints and demonstrate that, in most cases, even when only half of the extracted samples in \(\mathcal{D}_{poison}\) are poisoned, our approach can still use random sampling for \(\mathcal{D}_{clean}\) to effectively purify the model.}

\revision{Let the constant values $s_{asr}(M_v)$ and $s_{acc}(M_v)$ in~\cref{eqn:asr} and~\cref{eqn:acc} be denoted as $c_1$ and $c_2$, respectively, for simplicity. We use $\text{ASR}$ and $\text{ACC}$ to represent the ground-truth attack success rate and clean accuracy of a model (that can be measured on oracle poisoned and clean test sets), respectively.}

\revision{Assuming we have two imperfect proxy datasets: a proxy ``clean'' dataset that contains a proportion $\rho$ of poisoned data and a proxy ``poisoned'' dataset contains a proportion $1-\lambda$ of the non-poisoned data. Denote the poisoned and clean data distributions as $\mathcal{P}$ and $\mathcal{C}$, respectively.}

\revision{Then, by the rule of total probability and definition in~\cref{sec:preliminary}, we obtain:
\begin{align*}
    s_{acc}(M) & = \mathbb{E}_{x \sim \mathcal{D}_{clean}} [\mathbf{1} (M(x) = y)] \\
    & = p(x \in \mathcal{P} \mid x \sim \mathcal{D}_{clean}) \\
    &\quad \cdot p(M(x) = y \mid x \in \mathcal{P}, x \sim \mathcal{D}_{clean}) \\
    &\quad + p(x \in \mathcal{C} \mid x \sim \mathcal{D}_{clean}) \\
    &\quad \cdot p(M(x) = y \mid x \in \mathcal{C}, x \sim \mathcal{D}_{clean})\\
    & = \rho \text{ASR} + (1-\rho) \text{ACC}.
    % s_{asr}(M) &= \mathbb{E}_{x \sim \mathcal{D}_{poison}} [\mathbf{1} (M(x) = y)] \\
    % &= p(x \in \mathcal{P} \mid x \sim \mathcal{D}_{poison}) \cdot p(M(x) = y \mid x \in \mathcal{P}, x \sim \mathcal{D}_{poison}) \\
    % &\quad + p(x \in \mathcal{C} \mid x \sim \mathcal{D}_{poison}) \cdot p(M(x) = y \mid x \in \mathcal{C}, x \sim \mathcal{D}_{poison})\\
    % & = \lambda \text{ASR} + (1-\lambda) \text{ACC}
\end{align*}}

\revision{Similarly, for \( x \sim \mathcal{D}_{poison} \), we have:
\begin{align*}
    s_{asr}(M) & = \lambda \cdot \text{ASR} + (1-\lambda) \cdot \text{ACC}.
\end{align*}}

\revision{Substituting these into~\cref{eqn:objective} gives us $(1-\alpha)c_1 + \alpha(1-c_2) + \left( \alpha \rho + \alpha \lambda - \lambda \right) \text{ASR} + \left(2\alpha - 1 -\alpha \rho -\alpha\lambda + \lambda\right) \text{ACC}$. From this, the constraints ensuring effective purification are:
\begin{align}
    \left\{
    \begin{array}{ll}
        \alpha \rho < (1-\alpha) \lambda, \\
        \alpha (1-\rho) > (1-\alpha) (1-\lambda).
    \end{array}
    \right.
\end{align}}
\revision{That is, as long as the clean portion in \( \mathcal{D}_{clean} \) receives more attention than that in \( \mathcal{D}_{poison} \), our method will effectively remove backdoor modules while preserving utility; the same applies to the poisoned portion in reverse.}

\revision{To further illustrate, for an accurate proxy dataset, the constraints are simplified to $\alpha(1+\rho) - 1 < 0$ and $\alpha(1-\rho) > 0$. Thus, if $\rho < 1$, setting $\alpha$ such that $0 < \alpha < \frac{1}{\rho+1}$ can ensure effective purification.}

\revision{In our experiments, $\alpha$ is set to 0.4 by default. Then, the requirement becomes $\rho < \frac{3}{2}\lambda-\frac{1}{2}$. This implies that even with a random-guess backdoor data detection method, we can tolerate up to \( 25\% \) poisoned data in \(\mathcal{D}_{clean}\). Given that real-world poisoning rates are usually low (\eg below \( 20\% \) or even \( 1\% \)), random sampling for the ``clean'' proxy dataset remains highly feasible in practice.}

\section{AI Assistants}
We use ChatGPT/Gemini for writing and formatting supports, including grammar checks, improving the clarity of figure and table captions, and other surface-level edits.

\end{document}